\documentclass{article}

\PassOptionsToPackage{authoryear, round}{natbib}

\usepackage[final]{neurips_2025}

\usepackage[utf8]{inputenc} 
\usepackage[T1]{fontenc}    
\usepackage{url}            
\usepackage{booktabs}       
\usepackage{amsfonts}       
\usepackage{nicefrac}       
\usepackage{microtype}      
\usepackage{parskip}
\usepackage{xcolor}         
\usepackage{amsmath,amsthm}
\usepackage{graphicx}
\usepackage[T1]{fontenc}
\usepackage[utf8]{inputenc}
\usepackage[inline]{enumitem}
\usepackage{dsfont}
\usepackage{braket}

\usepackage[english]{babel}
\usepackage{amssymb,amsfonts}
\usepackage{bbm}
\usepackage{nicefrac}
\usepackage{xfrac}
\usepackage{subcaption}
\usepackage{booktabs}
\usepackage{multirow}
\usepackage{tikz}
\usetikzlibrary{arrows.meta}
\usepackage[colorinlistoftodos,bordercolor=purple,backgroundcolor=pink!20,linecolor=pink,textsize=scriptsize]{todonotes}
\usepackage{wrapfig}

\usepackage{thm-restate}
\usepackage{empheq}
\usepackage{color}
\definecolor{cool-blue}{HTML}{1D6996}
\definecolor{cool-purple}{HTML}{5F4690}
\definecolor{cool-teal}{HTML}{38A6A5}
\definecolor{cool-green}{HTML}{0F8554}
\definecolor{dark-green}{rgb}{0.0,0.5,0.0}

\newcommand{\algname}[1]{{\small\sf#1}}
\usepackage[pagebackref]{hyperref}
\hypersetup{
	colorlinks=true,
	linkcolor=cool-blue,
	filecolor=cool-blue,
	citecolor=dark-green,
	urlcolor=cool-blue
}
\renewcommand*{\backrefalt}[4]{%
    \ifcase #1 \footnotesize{(Not cited.)}%
    \or        \footnotesize{(Cited on page~#2)}%
    \else      \footnotesize{(Cited on pages~#2)}%
    \fi}

\usepackage{algorithm}
\usepackage{algpseudocode}
\usepackage[capitalize]{cleveref}
\usepackage[theorems,skins]{tcolorbox}
\theoremstyle{plain}
\newtheorem{theorem}{Theorem}[section]

\newtheorem{example}[theorem]{Example}
\newtheorem{lemma}[theorem]{Lemma}

\theoremstyle{definition}
\newtheorem{definition}[theorem]{Definition}
\newtheorem{assumption}[theorem]{Assumption}
\theoremstyle{remark}

\newcommand{\EE}{\mathbb{E}}

\newcommand{\cA}{\mathcal{A}}

\newcommand{\cS}{\mathcal{S}}
\newcommand{\cF}{\mathcal{F}}

\newcommand{\cM}{\mathcal{M}}

\newcommand{\R}{\mathbb{R}}

\newcommand{\Prob}{\mathbb{P}}

\newcommand{\1}{\mathds{1}}

\newcommand{\indic}[1]{\mathds{1}\left\{#1\right\}}

\newcommand{\iid}{\buildrel \text{i.i.d}\over\sim}

\def\<#1,#2>{\langle #1,#2\rangle}

\definecolor{lxs}{RGB}{138,43,226}

\NewDocumentCommand{\Var}{somo}{\mathrm{Var}\IfValueT{#2}{_{#2}}{} \IfBooleanTF{#1}{#3}{\IfValueTF{#4}{\!\left(#3\ \middle|\ #4\right)}{\parentheses*{#3}}}}

\newcommand{\polclass}{\Pi^{\text{s}}}
\newcommand{\realmdp}{\cM^\star}
\usepackage{ifthen}

\title{SPiDR: A Simple Approach for Zero-Shot Safety in Sim-to-Real Transfer
}

\author{%
    Yarden As\thanks{Corresponding author: \texttt{yardas@ethz.ch}} \\
    ETH Zurich \\
    \And
    Chengrui Qu \\
    Caltech \\
    \And
    Benjamin Unger \\
    ETH Zurich
    \And
    Dongho Kang \\
    ETH Zurich \\
    \And
    Max van der Hart \\
    ETH Zurich \\
    \And
    Laixi Shi \\
    Johns Hopkins University \\
    \And 
    Stelian Coros \\
    ETH Zurich \\
    \And
    Adam Wierman \\
    Caltech \\
    \And
    Andreas Krause \\ 
    ETH Zurich
}

\begin{document}

\maketitle

\begin{abstract}
\looseness=-1
Deploying reinforcement learning (RL) safely in the real world is challenging, as policies trained in simulators must face the inevitable `sim-to-real gap'. Robust safe RL techniques are provably safe however difficult to scale, while domain randomization is more practical yet prone to unsafe behaviors.
We address this gap by proposing \algname{SPiDR}, short for \textbf{S}im-to-real via \textbf{P}ess\textbf{i}mistic \textbf{D}omain \textbf{R}andomization---a scalable algorithm with provable guarantees for safe sim-to-real transfer. \algname{SPiDR} uses domain randomization to incorporate the uncertainty about the sim-to-real gap into the safety constraints, making it versatile and highly compatible with existing training pipelines. Through extensive experiments on sim-to-sim benchmarks and two distinct real-world robotic platforms, we demonstrate that \algname{SPiDR} effectively ensures safety despite the sim-to-real gap while maintaining strong performance.
\end{abstract}

\vspace{0.25cm}
\section{Introduction}
\label{sec:introduction}
Reinforcement learning (RL) has made significant strides in recent years, demonstrating remarkable progress across a range of domains. These include achieving superhuman capabilities in games~\citep{mnih2015humanlevel,44806}, fine-tuning large language models \citep{ouyang2022traininglanguagemodelsfollow}, advancing applications in healthcare \citep{pmlr-v126-fox20a,s20185058}, robotics \citep{joonho2020,degrave2022magnetic,lin2025sim} and autonomous driving \citep{cusumano2025robust,cornelisse2025building}. 
Yet despite these achievements, ensuring safety and preventing harmful behaviors remains a critical challenge and a prerequisite for unlocking the full potential of RL as a ubiquitous element in everyday life \citep{amodei2016concrete,gu2022review}.

The use of simulators has been a key component behind the success of many of the mentioned applications ~\citep{visentin2014university,makoviychuk2021isaac,degrave2022magnetic,kazemkhani2024gpudrive}. Training in simulation allows agents to learn from unsafe interactions, 
which in reality would lead to catastrophic outcomes. In addition, learning complex behaviors fully online can be prohibitively time-consuming. Modern simulators accelerate training, reducing hours of real-world experience to minutes on consumer-grade GPUs~\citep{pmlr-v164-rudin22a}. However, while being a major driver in the development of the above examples, even state-of-the-art simulators often fall short in precisely mirroring the real-world. Indeed, ``all models are wrong'' \citep{box1976science}---the so-called \emph{sim-to-real gap} can make simulation-trained policies violate real-world constraints, which can be particularly dangerous in high-stakes settings where safety must be guaranteed on first contact.

Existing literature to address this challenge often relies on tools from robust optimization~\citep{risk-averse,sota-rcmdp,zhang2024distributionally}. While being theoretically grounded, such methods typically require practitioners to significantly alter their existing training pipelines, rendering them less prevalent in practice. In contrast, due to its simplicity, domain randomization has become the de facto tool for sim-to-real transfer~\citep{tobin2017domain,peng2018sim2real,joonho2020,degrave2022magnetic}. Despite its success, in problems that require adherence to safety constraints, domain randomization lacks safety guarantees and often fails to satisfy the constraints in practice~\citep[cf.][and \Cref{fig:real-world-safety}]{risk-averse}. Therefore, a method that provably guarantees safe sim-to-real transfer, while being highly compatible with standard training practices, is still missing.

\begin{figure}
    \centering
    \includegraphics{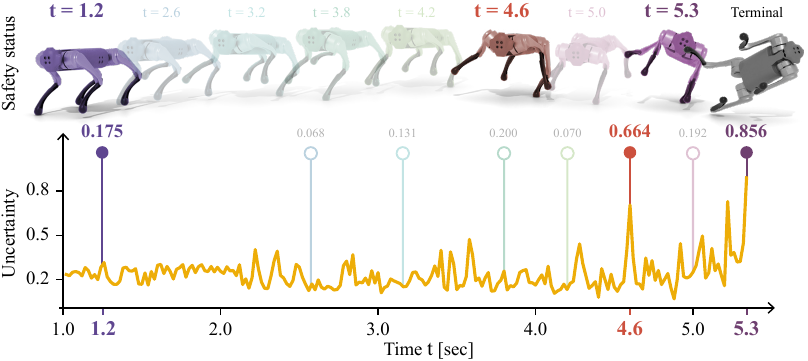}
    \caption{Uncertainty over a quadruped robot’s trajectory. The snapshots illustrate the robot's pose at key moments, with corresponding uncertainty levels highlighted. High-uncertainty transitions are incorporated into the cost function to discourage the policy from entering regions where the simulator is inaccurate and behavior is more likely to become unsafe during real-world deployment.}
    \label{fig:disagreement-demo}
\end{figure}

\looseness=-1In this work, we address this gap by presenting a simple method that builds on domain randomization while ensuring safety under sim-to-real transfer. We theoretically show that unsafe transfer can be associated with large \emph{uncertainty} about the sim-to-real gap, quantified as the disagreement among next-state predictions from domain-randomized dynamics models. This key idea is illustrated in \Cref{fig:disagreement-demo}, where spikes in uncertainty (e.g. at $t = 4.6$ and $t = 5.3$) coincide with unstable or unsafe behaviors, such as stumbling or flipping. Motivated by this insight, we propose to penalize the cost with the uncertainty to achieve safe sim-to-real transfer, leading to the design of \algname{SPiDR}. Notably, \algname{SPiDR} seamlessly integrates with state-of-the-art RL algorithms~\citep{schulman2017proximal,haarnoja2019soft}, delivering strong empirical performance on both in simulated and real-world safe RL tasks while ensuring constraint satisfaction, even under severe model mismatch. 
\paragraph{Out contribution.}
\begin{itemize}[leftmargin=0.5cm, parsep=0.3em]
    \item We address an important challenge to real-world adoption of RL: zero-shot safe sim-to-real transfer, where an agent must learn a safe and effective policy using only simulated interactions. We propose \algname{SPiDR}, a practical algorithm with formal safety guarantees that integrates easily into popular sim-to-real pipelines.
    \item We validate \algname{SPiDR} on two real-world robotic platforms, where it achieves zero-shot constraint satisfaction, substantially outperforming other baselines in terms of safety and performance. These results provide empirical evidence that our theoretical guarantees translate to the real-world, suggesting that \algname{SPiDR} can be safely used in real-world deployment.
    \item Finally, we extensively evaluate \algname{SPiDR} on well established simulated continuous control benchmarks, including the RWRL benchmark \citep{dulacarnold2020realworldrlempirical}, Safety Gym \citep{Ray2019} and RaceCar environments \citep{kabzan2020amz}, where \algname{SPiDR} consistently satisfies safety constraints while achieving strong task performance.
\end{itemize}

\section{Related Works}
\label{sec:related-works}
Safe sim-to-real transfer can be naturally framed as a constrained Markov decision process \citep[CMDP,][]{altman-constrainedMDP} under model uncertainty. A common approach is to extend CMDPs using tools from robust optimization~\citep{iyengar2005robust,ben2009robust}, which has led to a growing body of work spanning both theoretical and practical contributions. We refer to \Cref{sec:additional-related-works} for a more comprehensive discussion of each line of work.

\paragraph{Provably robust algorithms.}
\looseness=-1
\citet{zhang2024distributionally} build on a game-theoretic formulation to develop a tractable primal-dual algorithm with provable non-asymptotic convergence guarantees to a safe policy. \citet{sota-rcmdp} take this further by proposing a policy gradient algorithm with formal guarantees for safety \emph{and} optimality, via an epigraph form of the robust CMDP problem. Robust CMDPs are inherently challenging since the worst-case scenarios w.r.t. the reward and cost can differ~\citep[see][]{sota-rcmdp}; both works make notable theoretical progress on this front. While sharing the goal of provably safe transfer, our work adopts a more scalable and modular approach by building on domain randomization, and integrating with existing CMDP solvers, avoiding the complexity of solving the minimax formulation, common in robust optimization.

\paragraph{Scalable robust algorithms.}
\looseness=-1\citet{russel2020robust} and \citet{mankowitz2020robust} were among the first to study robust CMDPs in the context of deep RL, proposing practical methods that scale to continuous control tasks. Similarly, \citet{risk-averse} introduce \algname{RAMU}, an algorithm that uses coherent risk measures~\citep{shapiro2017distributionally} with temporal differences (TD) learning, achieving strong empirical performance on the RWRL benchmark, though lacking formal safety guarantees.
\citet{sun2024constrained} extends CPO \citep{achiam2017constrained} to problems with model uncertainty, providing safety and performance guarantees. Lastly, \citet{bossens2024robust} proposes to learn an adversary policy and show that their algorithm is robust w.r.t. $L_1$-norm uncertainty sets. Our work differs in that it provides safety guarantees while remaining scalable and not tied to a particular RL algorithm.

\paragraph{Practical methods for safe sim-to-real.}
While the previous works develop methods for solving robust CMDPs---often motivated by the practical problem of safe sim-to-real transfer---other prior works address this problem directly. \citet{kaushik2022safeapt} use online data from the real system and employ safe Bayesian optimization \citep{pmlr-v37-sui15} to select safe and high-performing policies from a collection of policies trained in simulation. Similarly, \citet{hsu2023sim} propose a multi-fidelity approach~\citep{cutler2014reinforcement}, incorporating a fine-tuning step in a high-fidelity simulator prior to deployment. Both works emphasize practical applicability and provide strong empirical validation in real robotics settings. Compared to these works, in this work we focus on guaranteeing safe transfer without access to online data or a computationally expensive intermediate simulator.

\section{Problem Setting}
\label{sec:problem-setting}
\paragraph{Constrained Markov decision process.} 
\looseness=-1We study discounted infinite-horizon CMDPs, defined by a tuple $\cM = (\cS, \cA, p, r, c, \gamma, \rho)$. Here, $\cS$ and $\cA$ are the state and action spaces, $p:\cS\times \cA\rightarrow \Delta(\cS)$ denotes the transition probability of the system dynamics, the reward function is given by $r:\cS\times \cA\rightarrow [0, r_{\text{max}}]$ and the cost is given by $c:\cS\times \cA\rightarrow [0, c_{\text{max}}]$. 
The discount factor is $\gamma \in [0, 1)$ and $\rho \in \Delta(\cS)$ is a probability distribution from which initial states are drawn. We consider the class of stationary policies $\polclass$, where each policy is a stochastic mapping from states to actions $\pi:\cS\rightarrow \Delta(\cA).$\footnote{While only the class of history-dependent policies is formally complete under domain randomization, we focus on $\polclass$ due to its simplicity. Our theoretical and empirical results can be directly extended to history-dependent policies. See \citet{dolgov2005stationary,kwon2021rl,chen2022understandingdomainrandomizationsimtoreal} for further discussions.} For given dynamics $p$, the value function under policy $\pi$ at state $s$ is defined as $V_r^{p,\pi}(s)\triangleq\EE_{p,\pi}[\sum_{t=0}^\infty\gamma^tr_t \mid s]$. Similarly, the cost value is given by $V_c^{p,\pi}(s)\triangleq\EE_{p,\pi}[\sum_{t=0}^\infty\gamma^tc_t \mid s]$. We define the expected value function of $\pi$ when the initial state is sampled from $\rho$ as $J_p(\pi)\triangleq\EE_{s\sim \rho}[V_r^{p,\pi}(s)]$, and the expected cost value as $C_p(\pi)\triangleq\EE_{s\sim \rho}[V_c^{p,\pi}(s)]$. The goal is to find a policy $\pi$ that solves
\begin{equation}
    \max_{\pi \in \polclass} J_p(\pi) \quad \text{s.t.} \quad C_p(\pi) \le d,
\end{equation}
where $d>0$ is a predefined budget. This formulation enables explicit decoupling of safety from the objective. For example, in robotics, the cost can represent collisions with obstacles, while the reward encourages reaching a goal.

\paragraph{Safe sim-to-real transfer.} In this work, we consider a setting where the agent has access to a simulator, capable of generating any environment $\widehat{\cM}_\xi = (\cS, \cA, \hat{p}_\xi, r, c, \gamma, \rho)$ given $\xi \in \Xi \subseteq \R^{d_\xi}$. The dynamics of each environment are parameterized by $\xi$, typically representing the physical properties of the system dynamics. The agent can freely interact with any simulated environment $\hat{p}_\xi$ with $\xi \in \Xi$, but has no access to the unknown real environment $\realmdp=(\cS, \cA, p^\star, r, c, \gamma, \rho)$. The objective is to learn a policy $\pi \in \polclass$ purely within the simulator such that, when deployed in the real environment $\realmdp$, it satisfies the safety constraint $C_{p^\star}(\pi) \le d$. Crucially, the agent must guarantee constraint satisfaction ``zero-shot'', without any direct interaction with the real environment $\realmdp$.

\section{{\sf SPiDR} for Safe Zero-Shot Sim-to-Real Transfer}
\label{sec:method}

\subsection{Domain Randomization}
\paragraph{Extending domain randomization to CMDPs.}
\looseness=-1
Domain randomization is particularly well-suited for the problem setting described above, as it leverages a set $\{\xi_i\}_{i=1}^N \iid \mu$ of parameterized environments, sampled independently from some probability distribution $\mu$. This distribution acts as a \emph{prior} for the real, yet unknown system parameters. 
A natural approach for tackling safe sim-to-real problems is by formulating CMDPs over a distribution of randomized domains, i.e., solving
\begin{equation}\label{eq:DR-average}
    \max_{\pi \in \polclass} \EE_{\xi\sim \mu}J_{\hat{p}_\xi}(\pi) \quad \text{s.t.} \quad \EE_{\xi\sim \mu}C_{\hat{p}_\xi}(\pi) \le d.
\end{equation}
Domain randomization can be seen as a sample average approximation of \Cref{eq:DR-average}, making it both straightforward to implement and scalable, as massively-parallel simulators can be used to collect data from each environment in parallel.

\paragraph{Domain randomization is not always safe.}
While \Cref{eq:DR-average} provides a compelling formulation from a practical standpoint, it does not guarantee safety in the real environment $\realmdp$. Specifically, since simulators only approximate the real world with limited precision, as well as due to averaging over dynamics, the costs in \Cref{eq:DR-average} may \emph{underestimate} the true costs in $\realmdp$. This limitation is empirically validated in \Cref{sec:real-world,sec:sim-to-sim}~(\Cref{fig:real-world-safety,fig:simulated-costs}, respectively) and theoretically illustrated through an example in \cref{sec:dr-failure-modes}. In what follows, we formally characterize constraint underestimation and show how \algname{SPiDR} is designed to mitigate it.

\paragraph{A pessimistic upper bound.}
\looseness=-1We quantify the extent by which constraints on the real system may be underestimated by establishing the following bound. To this end, we measure the discrepancy between the simulated and real dynamics using the $L_1$-Wasserstein distance, denoted as $D_W(\hat{p}_\xi, p^\star)(s,a)$, whose formal definition is provided in \cref{definition:wasserstein}. While our analysis is based on this metric, it naturally extends to other discrepancy measures. We assume this discrepancy is finite for all $\xi \in \Xi$, which is a reasonable assumption in practice. For instance, the simulators by \citet{makoviychuk2021isaac} and \citet{zakka2025mujocoplayground} have been successfully used for zero-shot sim-to-real transfer across several robotic platforms, suggesting that they maintain high fidelity with real-world systems. We now present our bound below.
\begin{lemma}\label{lem:simulation}
    Let $\Prob_{p,\pi,t}(s)$ denote the probability of reaching the state $s$ at step $t$ under the policy $\pi$ and the dynamics $p$, and let $d_{p,\pi}\triangleq(1-\gamma)\pi(a|s)\sum_{t=0}^{\infty}\gamma^t \Prob_{p,\pi,t}(s)$ denote the normalized discounted occupancy measure of policy $\pi$ under the dynamics $p$. The real-world cost $C_{p^\star}(\pi)$ can be upper-bounded by
    \begin{align}\label{eq:cost-underestimation}
        C_{p^\star}(\pi) &\le \underbrace{\EE_{\xi\sim \mu}C_{\hat{p}_\xi}(\pi)}_{\text{Constraint in simulation}} + \EE_{\xi\sim \mu} \Big[\EE_{(s,a)\sim d_{\hat{p}_\xi,\pi}} \Big[\frac{\gamma L_C}{1 - \gamma}D_W(\hat{p}_\xi, p^\star)(s, a) \Big] \Big],
    \end{align}
    where $L_C$ is the Lipschitz constant of the state cost function $V_c^{p^\star,\pi}(s)$.
\end{lemma}
\Cref{lem:simulation} shows that the true safety constraint function $C_{p^\star}(\pi)$ is upper-bounded by the constraint evaluated during training, and the expected $L_1$-Wasserstein distance with respect to the state-action occupancy measure of $\pi$ over the simulated dynamics $\hat{p}_\xi$. Importantly, it establishes that even when the constraint is satisfied in simulation, i.e., $\EE_{\xi\sim \mu}C_{\hat{p}_\xi}(\pi) \le d$, the constraint on the real system may be larger, depending on $\pi$ and the degree of mismatch between the simulated and real dynamics. Therefore, by bounding the r.h.s. of \Cref{eq:cost-underestimation} with $d$, we guarantee that $C_{p^\star}(\pi) \le d$, hence safe transfer to the real system. We refer to \Cref{sec:proofs} for the formal proof and assumptions of \Cref{lem:simulation}. We next show how this key insight is used in our design of \algname{SPiDR}.

\subsection{Algorithm Design}
\paragraph{Reduction to penalized CMDPs.}
\looseness=-1 Observing that by linearity of expectation, the r.h.s. of \Cref{eq:cost-underestimation} can be written as 
\begin{equation*}
    \EE_{\xi\sim \mu} \left[\EE_{(s,a)\sim d_{\hat{p}_\xi,\pi}}\left[c(s, a) + \frac{\gamma L_C}{1 - \gamma}D_W(\hat{p}_\xi, p^\star)(s, a)\right]\right].
\end{equation*}
This simple insight motivates the use of
\begin{equation}
    \tilde{c}(s, a) \triangleq c(s, a) + \underbrace{\frac{\gamma L_C}{1 - \gamma} \max_{\xi\in\Xi}D_W(\hat{p}_\xi, p^\star)(s, a)}_{\text{penalty}} \label{eq:penalty-term}
\end{equation}
as a surrogate cost function during training. Crucially, it is infeasible to estimate the model discrepancy $D_W(\hat{p}_\xi, p^\star)(s, a)$ for any $\xi\in\Xi$, therefore we use the worst-case one as a conservative approximation. Using $\tilde{c}(\cdot, \cdot)$ yields a \emph{penalized} CMDP $(\cS, \cA, \hat{p}_\xi, r, \tilde{c}, \gamma, \rho)$. This CMDP is still fully compatible with domain randomization, allowing us to solve 
\begin{equation}
    \label{eq:pessimistic-cmdp}
    \max_{\pi\in\polclass}\EE_{\xi\sim \mu}J_{\hat{p}_\xi}(\pi) \quad \text{s.t.} \quad \EE_{\xi\sim \mu}\widetilde{C}_{\hat{p}_\xi}(\pi) \le d,
\end{equation}
where $\widetilde{C}_{\hat{p}_\xi}(\pi)$ denotes the constraint with $\tilde{c}(\cdot, \cdot)$ following $\hat{p}_\xi$. While $\tilde{c}(\cdot, \cdot)$ can be used as a conservative approximation that in principle guarantees safe transfer, direct access to the penalty term is generally intractable. This is due to the fact that $L_C$ and $D_W(\cdot, \cdot)$ are unknown a priori and can only be estimated using access to ground-truth data from the real system. Therefore, we propose approximating the penalty term using only simulated data.

\paragraph{Approximating the penalty term.}
\looseness=-1We propose using an ensemble of $\{\hat{p}_{\xi_i}\}_{i = 1}^n \iid \mu$ dynamics and measuring their disagreement in predicting the next state, as a proxy for the uncertainty about the model discrepancy $\max_{\xi \in \Xi}D_W(\hat{p}_\xi,p^\star)$. Specifically, when $s_i \sim \hat{p}_{\xi_i}(\cdot \mid s, a)$ are $n$ i.i.d. samples,  we define the sum of component-wise empirical variances of the ensemble next-state predictions as the estimator
\begin{equation}
\label{eq:variance-estimator}
    \upsilon(s, a) \triangleq \left\| \mathrm{Var} \left( s_1,\ldots,s_n \right) \right\|_1 = \sum_{j=1}^{\operatorname{dim}(\cS)} \mathrm{Var} \left( s_{1,j},\ldots,s_{n,j} \right)
\end{equation}
where $\operatorname{dim}(\cS)$ is the dimension of the state space and $s_{i,j}$ denotes the $j$-th component of the $i$-th sample. The empirical variance $\upsilon(s, a)$ measures the sensitivity of the environment in predicting the next state w.r.t. $\xi$, making it an effective proxy for our uncertainty about the model discrepancy $\max_{\xi \in \Xi}D_W(\hat{p}_\xi,p^\star)$, especially when the real environment lies near the simulated ones. In \Cref{sec:heuristic}, we formally show that for a suitable constant $\lambda$, and under bounded model mismatch, $\lambda \upsilon(s, a)$ upper-bounds $\frac{\gamma L_c}{1 - \gamma}\max_{\xi \in \Xi} D_W(\hat{p}_\xi, p^\star)(s, a)$, with the bound becoming tighter as $n$ increases. Additional practical guidance on how to pick $\lambda$ empirically is provided in \Cref{sec:choosing-lambda}. We find this approach simple to implement, computationally efficient and effective in practice, as demonstrated in \Cref{sec:experiments}~(\Cref{fig:simulated,fig:real-world-safety}). Moreover, in \Cref{fig:scaling} we demonstrate how to pick $n$ in practice.

\paragraph{The algorithm.}
With the above approximation $\upsilon(\cdot, \cdot)$ in hand, we are ready to introduce the entire algorithm \algname{SPiDR}, summarized in \Cref{alg:spidr}.
Standard domain randomization typically involves policy search methods such as policy gradients or TD learning, which require collecting trajectory data from the simulator. These trajectories are collected independently and in parallel for each dynamics $\{\xi_i\}_{i = 1}^{N}$. To incorporate pessimism, we modify only the procedure by which these trajectories are obtained. This abstraction allows us to keep using domain randomization as it is while remaining versatile w.r.t. the choice of policy search algorithm.
\begin{algorithm*}[h]
\caption{\algname{SPiDR}: Safe Sim-to-Real via Pessimistic Domain Randomization}
\algrenewcommand\algorithmicindent{1em}%
\label{alg:spidr}
\begin{algorithmic}[1]
\State \textbf{Input:} pessimism $\lambda$, initial distribution $\mu$, behavior policy $\pi$
\State \textbf{Init:}  Sample $\{\xi_i\}_{i = 1}^N \times \{\xi_{ij}\}_{j = 1}^n \iid \mu$ \label{alg:draw-dynamics}
\For{$i = 1, \dots, N$ \textbf{in parallel}} \label{alg:parallel-traj}
    \Comment{Collect data from each $\xi_i$}
    \State Initialize trajectory $\tau^{(i)} \gets \emptyset$
    \For{$t = 0, 1, \dots$}
        \Comment{Rollout policy}
        \State $a_t \sim \pi(\cdot \mid s_t)$ 
        \State Simulate $s_{t+1} \sim \hat{p}_{\xi_i}(\cdot \mid s_t, a_t)$, obtain $r_t$, $c_t$ \label{alg:propagate}
        \State \label{alg:one-step}Simulate $s_{t + 1}^{(j)} \sim \hat{p}_{\xi_{ij}}(\cdot \mid s_t, a_t)$ with each $\{\hat{p}_{\xi_{ij}}\}_{j = 1}^n$
        \Comment{Fast parallel execution}
        \State \label{alg:penalty}Compute the penalty term $\upsilon(s_t, a_t)$ in \Cref{eq:variance-estimator}
        \State \label{alg:cost}Penalize cost $\tilde{c}_t \gets c(s_t, a_t) + \lambda \upsilon(s_t, a_t)$
        \State Append $(s_t, a_t, r_t, \tilde{c}_t)$ to $\tau^{(i)}$
        \State Set next state $s_t \gets s_{t + 1}$ \label{alg:any-cmdp}
    \EndFor
\EndFor
\State Solve \Cref{eq:pessimistic-cmdp} using $\{\tau^{(i)}\}_{i=1}^N$ to obtain $\tilde{\pi}$ with any CMDP solver
\State \Return $\tilde{\pi}$
\end{algorithmic}
\end{algorithm*}
Specifically, following standard domain randomization, \algname{SPiDR} samples a batch of dynamics $\{\xi_i\}_{i = 1}^N$. These dynamics are used only to rollout the policy~(\Cref{alg:draw-dynamics,alg:propagate}). For each sampled dynamics, we further employ an ensemble $\{\hat{p}_{\xi_{ij}}\}_{j = 1}^n$. This ensemble is used to estimate the penalty term as $\upsilon(\cdot, \cdot)$ (\Cref{alg:one-step,alg:penalty,alg:cost}). These trajectories in the ``penalized'' CMDP using $\tilde{c}$ are collected and used to solve the penalized CMDP in \Cref{eq:pessimistic-cmdp}. Importantly, \Cref{alg:one-step,alg:penalty,alg:cost} represent the only modifications made relative to standard domain randomization.

\subsection{Safety Guarantee}\label{sec:theoretical-things}
Next, we demonstrate our theoretical guarantees for a solution to \Cref{eq:pessimistic-cmdp}. We first assume the feasibility of the problem, otherwise, a solution can be recovered only by improving the simulator or with safe online learning techniques \citep{as2025actsafe}. We show that any solution to the penalized CMDP provably achieves safety in the real environment by the following theorem.

\begin{theorem}\label{thm:safety-performance}
Let $p^\star$ be the dynamics of the real environment $\realmdp$. Let $\tilde{\pi}$ be a solution to the penalized CMDP introduced in \Cref{eq:pessimistic-cmdp}, i.e., $\tilde \pi = \max_{\pi\in\polclass}\EE_{\xi\sim \mu}J_{\hat{p}_\xi}(\pi) \text{ s.t. } \EE_{\xi\sim \mu}\widetilde{C}_{\hat{p}_\xi}(\pi) \le d$. Then $\tilde{\pi}$ satisfies the safety constraint in the real environment, namely, $C_{p^\star}(\tilde{\pi}) \leq d$. \label{eq:safety}
\end{theorem}
The formal proof of this theorem, including its assumptions, is provided in \Cref{sec:proofs}. Solving the CMDP with any penalty on the cost that dominates $\max_{\xi\in\Xi}D_W(\hat{p}_\xi, p^\star)$ yields a more conservative policy. Any such policy will remain safe in the real environment. The drop in performance due to cost penalization is not captured by \Cref{thm:safety-performance} as it also depends on cost and reward function. However, our real-world experiments in \Cref{sec:experiments}~(\Cref{fig:real-world-safety}) demonstrate that \algname{SPiDR} consistently matches or even outperforms standard domain randomization. Adopting any penalty on the cost reduces the set of feasible policies. Therefore, the agent may avoid high-reward regions that are in reality safe but uncertain in simulation. However, the limited drop observed in practice suggests that our bound $\upsilon(\cdot,\cdot)$ provides a reasonably tight over-approximation of the worst-case model discrepancy.

\section{Experiments}
\label{sec:experiments}
Next, we demonstrate \algname{SPiDR}'s performance in practice. Below we provide our sim-to-real experiments on two robotic tasks, followed by comprehensive ablations on several tasks from three well-established safe RL benchmarks. 
We refer the reader to \Cref{sec:additional-ablations,sec:choosing-lambda} for additional details on how $\lambda$ is picked in practice and for more ablations.

\paragraph{Setup.}
Unless otherwise specified, all experiments use SAC \citep{haarnoja2019soft,nauman2024bigger} in combination with either CRPO \citep{xu2021crpo} or a simple primal-dual constrained optimization method~\citep{bertsekas2016nonlinear}. We use these CMDP solvers since they deliver strong results while remaining easy to implement. We run each experiment with five random seeds and report the mean and standard error. Empirical estimates of the objective and constraint on the \emph{test} environments are denoted by $\hat{J}(\tilde{\pi})$ and $\hat{C}(\tilde{\pi})$, respectively. We compare \algname{SPiDR} with the following baselines: 
\begin{enumerate*}[label=\textbf{(\roman*)}]
    \item \algname{Nominal} is a simple baseline that collects trajectories only from the nominal training dynamics;
    \item \algname{Domain Randomization} collects trajectories only from the perturbed dynamics of the training distribution; and
    \item \algname{RAMU} \citep{risk-averse}, a state-of-the-art robust safe RL algorithm designed specifically for TD learning methods.
\end{enumerate*}

\subsection{Real-World Deployment}
\label{sec:real-world}
We demonstrate \algname{SPiDR}'s effectiveness on two real-world robotic tasks: a highly-dynamic remote-controlled race car, and on a Unitree Go1 quadruped robot, illustrated in \Cref{fig:hardware-demo}. Each policy is trained only in simulation and then evaluated in the real world: five trials for the remote-controlled car and ten trials for Unitree Go1. These trials are used to obtain $\hat{C}(\tilde{\pi})$ and $\hat{J}(\tilde{\pi})$ where applicable. In both tasks, we compare \algname{SPiDR} with domain randomization evaluated on the real system. Additionally, to demonstrate how domain randomization may underestimate the constraint in real, we compare its performance with \algname{SPiDR}, when evaluated \emph{during training in simulation}. Further details about the tasks and hardware are provided in \Cref{sec:racecar-real,sec:go1-locomotion}.
\newcommand{\elevation}{-10}
\newcommand{\anglerot}{77}

\begin{figure}
    \centering
    \begin{tabular}{@{\hskip 5pt} c @{\hskip 5pt} c @{\hskip 5pt} c @{\hskip 5pt} c @{\hskip 5pt}}
        \begin{tikzpicture}
            \node[anchor=south west, inner sep=0] (image1) at (0, 0) {\includegraphics[height=75pt, trim={16cm 10cm 13.75cm 3cm}, clip]{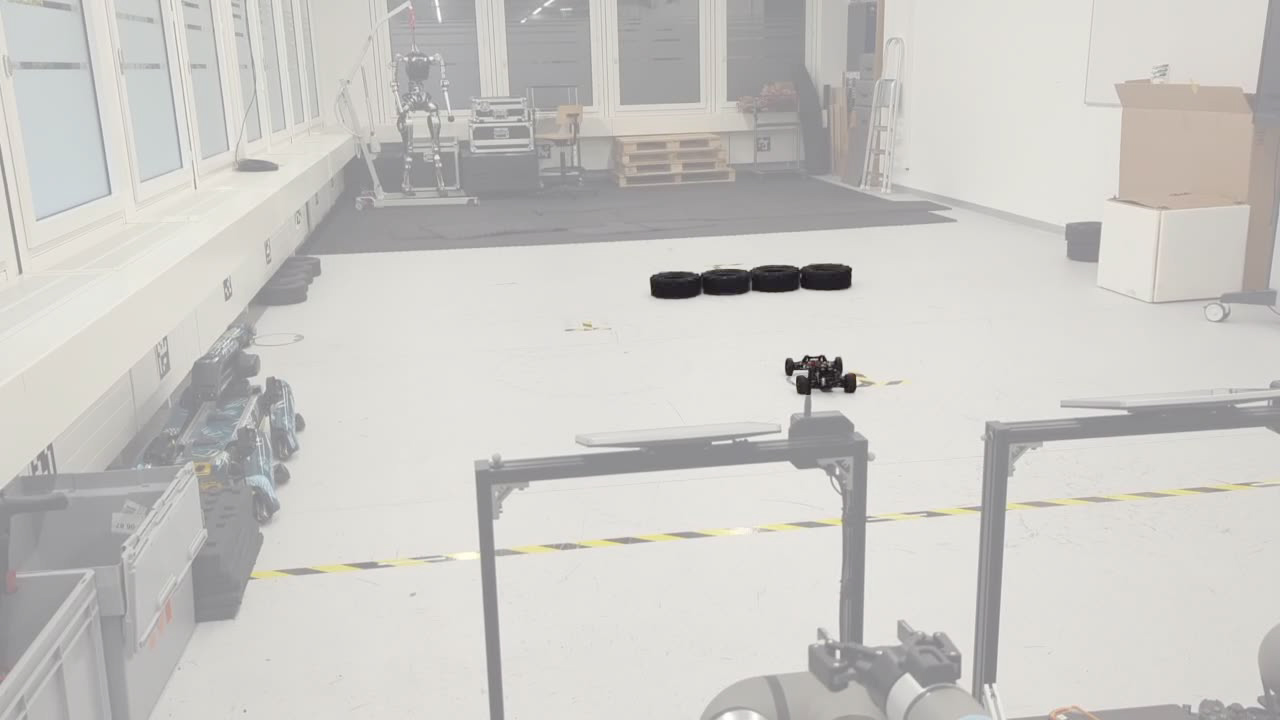}};
            \fill[red, opacity=0.5,xslant=-tan(\elevation+\anglerot), rotate=-\elevation, yscale=cos(\anglerot)] (5.25, 1.7) circle (7pt);
            \node[anchor=south west, fill=white, font=\small, text=black, rounded corners, xshift=1.64cm, yshift=2.cm] (goal) at (image1.south west) {Goal};
            \draw[-{Circle[open]}, thick] (goal.south) -- ++(0.0, -0.75);
            \node[anchor=south west, font=\bfseries\large] at (image1.south west) {t = 0[sec]};
        \end{tikzpicture} &
        \begin{tikzpicture}
            \node[anchor=south west, inner sep=0] (image2) at (0, 0) {\includegraphics[height=75pt, trim={16cm 10cm 13.75cm 3cm}, clip]{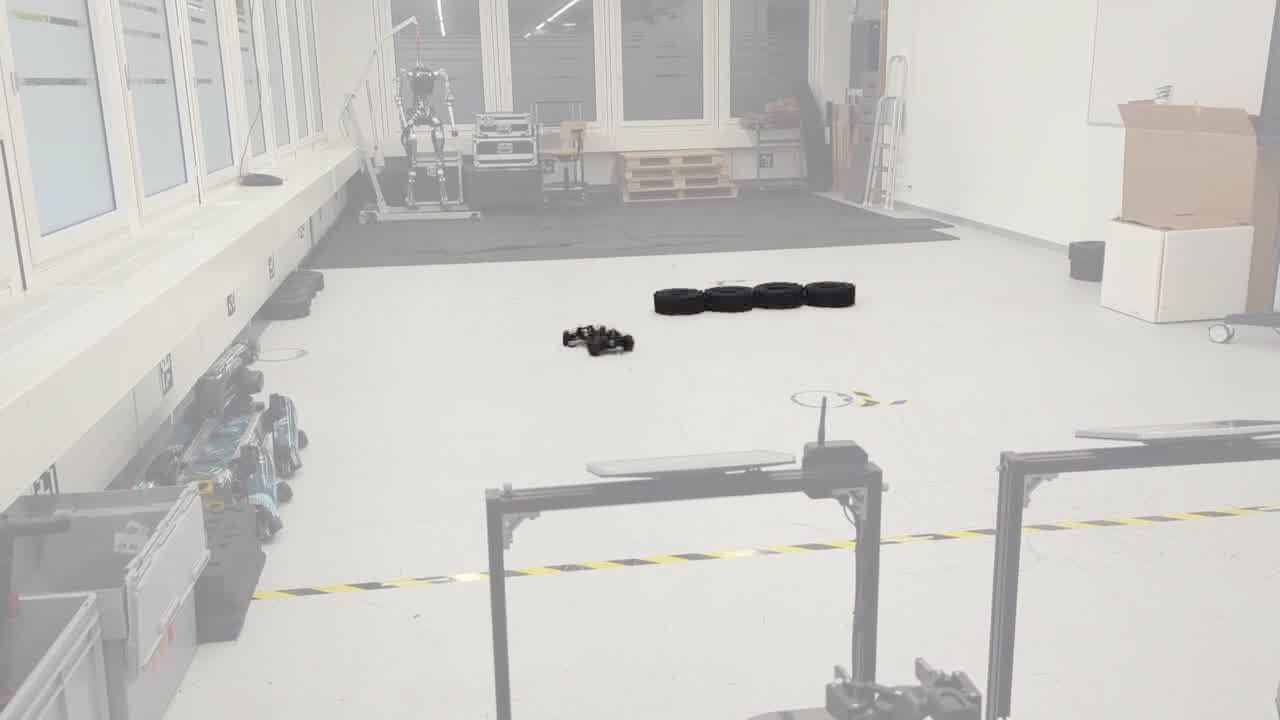}};
            \fill[red, opacity=0.5,xslant=-tan(\elevation+\anglerot), rotate=-\elevation, yscale=cos(\anglerot)] (5.1, 1.65) circle (7pt);
            \node[anchor=south west, fill=white, font=\small, text=black, rounded corners, xshift=1.64cm, yshift=2.cm] (obstacles) at (image1.south west) {Obstacles};
            \draw[-{Circle[open, color=white]}, thick] (obstacles.south) to[out=-90, in=90]  ++(-0.7, -1.04);
            \draw[-{Circle[open, color=white]}, thick] (obstacles.south) to[out=-90, in=90]  ++(-0.31, -1.01);
            \draw[-{Circle[open, color=white]}, thick] (obstacles.south) to[out=-90, in=90]  ++(0.04, -0.99);
            \draw[-{Circle[open, color=white]}, thick] (obstacles.south) to[out=-90, in=90]  ++(0.44, -0.97);
            \node[anchor=south west, font=\bfseries\large] at (image2.south west) {t = 3[sec]};
        \end{tikzpicture} &
        \begin{tikzpicture}
            \node[anchor=south west, inner sep=0] (image3) at (0, 0) {\includegraphics[height=75pt, trim={16cm 10cm 13.75cm 3cm}, clip]{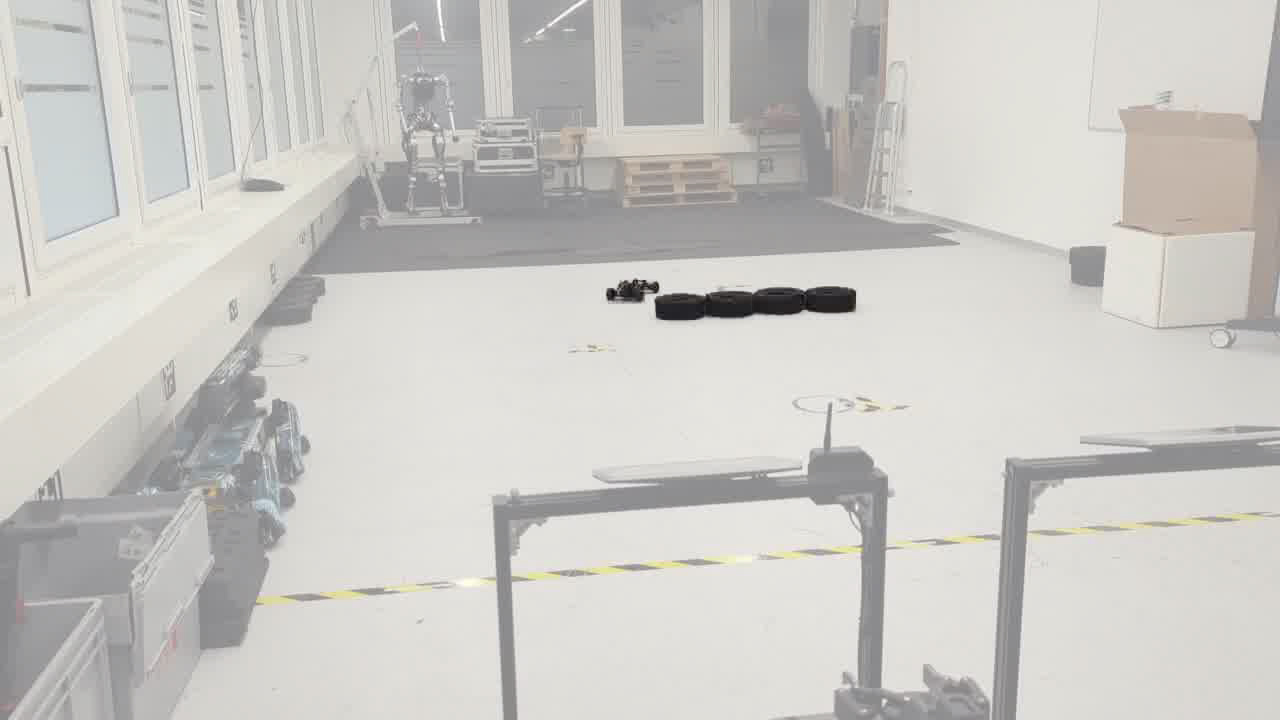}};
            \fill[red, opacity=0.5,xslant=-tan(\elevation+\anglerot), rotate=-\elevation, yscale=cos(\anglerot)] (5.05, 1.55) circle (7pt);
            \node[anchor=south west, font=\bfseries\large] at (image3.south west) {t = 6[sec]};
        \end{tikzpicture} &
        \begin{tikzpicture}
            \node[anchor=south west, inner sep=0] (image4) at (0, 0) {\includegraphics[height=75pt, trim={16cm 10cm 13.75cm 3cm}, clip]{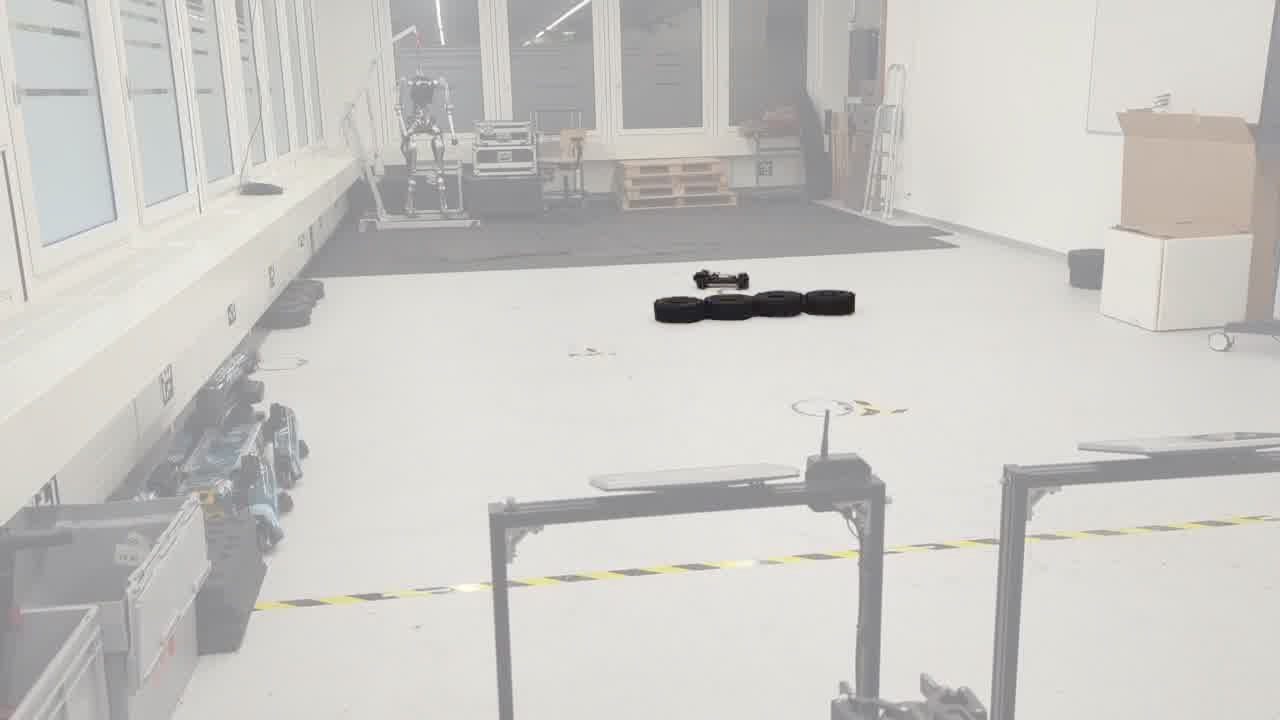}};
            \fill[red, opacity=0.5,xslant=-tan(\elevation+\anglerot), rotate=-\elevation, yscale=cos(\anglerot)] (5., 1.45) circle (7pt);
            \node[anchor=south west, font=\bfseries\large] at (image4.south west) {t = 7[sec]};
        \end{tikzpicture} \\
        \begin{tikzpicture}
            \node[anchor=south west, inner sep=0] (image1) at (0, 0) 
                {\includegraphics[width=0.233\textwidth, trim={14cm 0cm 0cm 0cm}, clip]{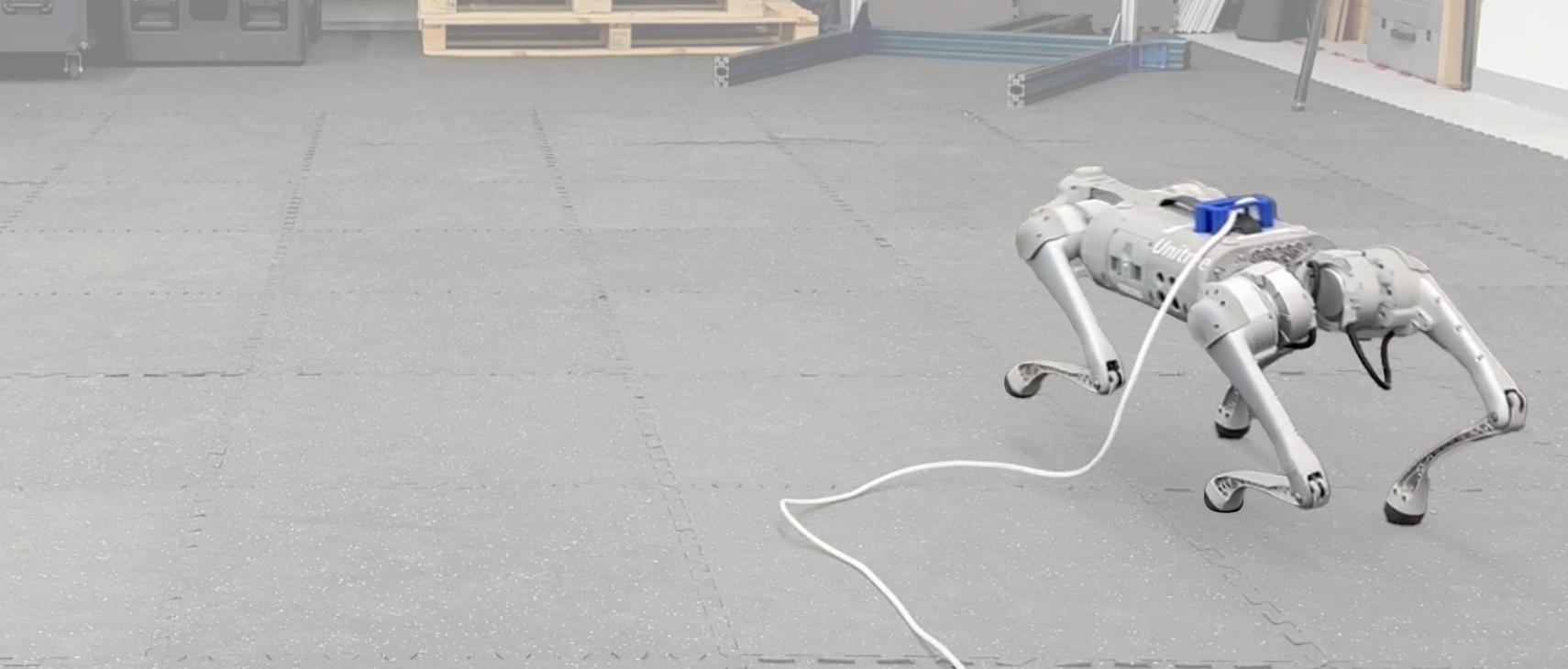}};
            \node[anchor=south west, font=\large\bfseries] at (image1.south west) {t = 0[sec]};
        \end{tikzpicture} &
        \begin{tikzpicture}
            \node[anchor=south west, inner sep=0] (image4) at (0, 0) 
                {\includegraphics[width=0.233\textwidth, trim={8cm 0cm 6cm 0cm}, clip]{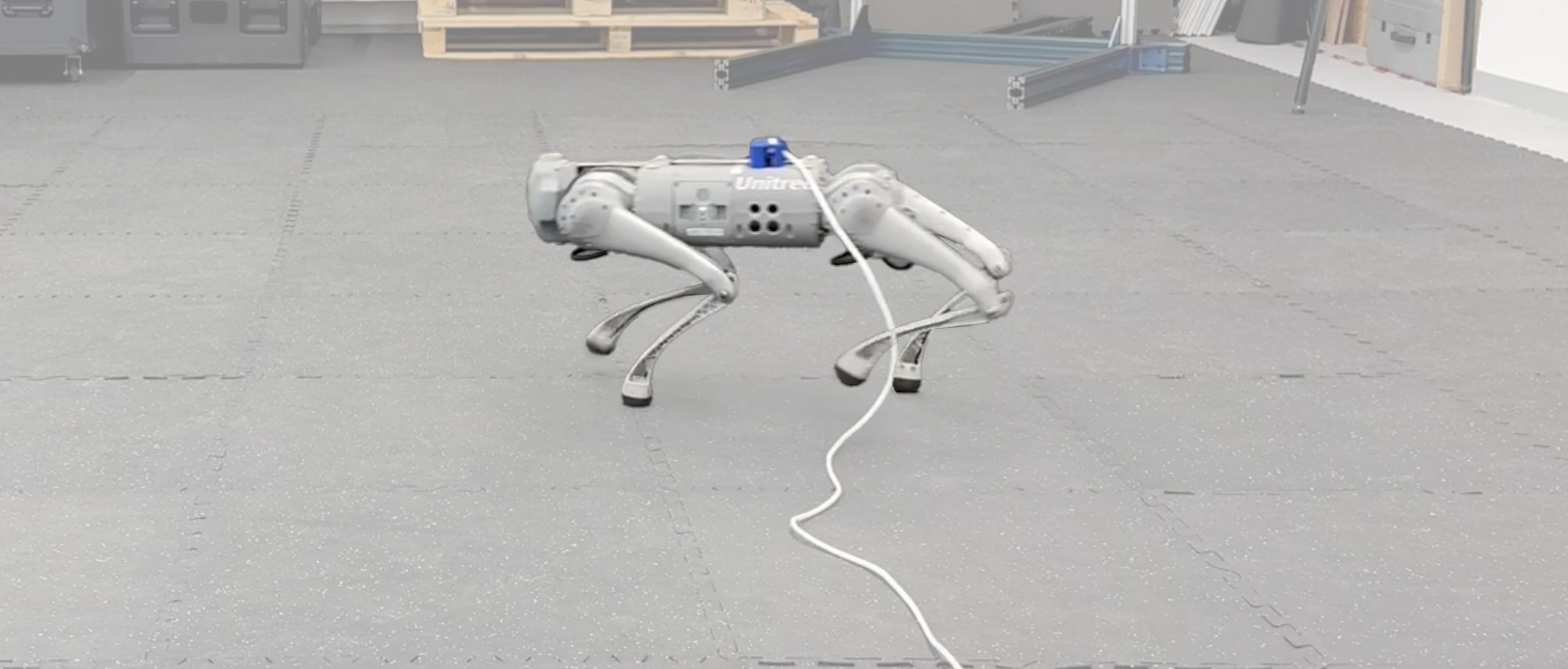}};
            \node[anchor=south west, font=\large\bfseries] at (image4.south west) {t = 1.5[sec]};
        \end{tikzpicture} &
        \begin{tikzpicture}
            \node[anchor=south west, inner sep=0] (image6) at (0, 0) 
                {\includegraphics[width=0.233\textwidth, trim={3cm 0cm 11cm 0cm}, clip]{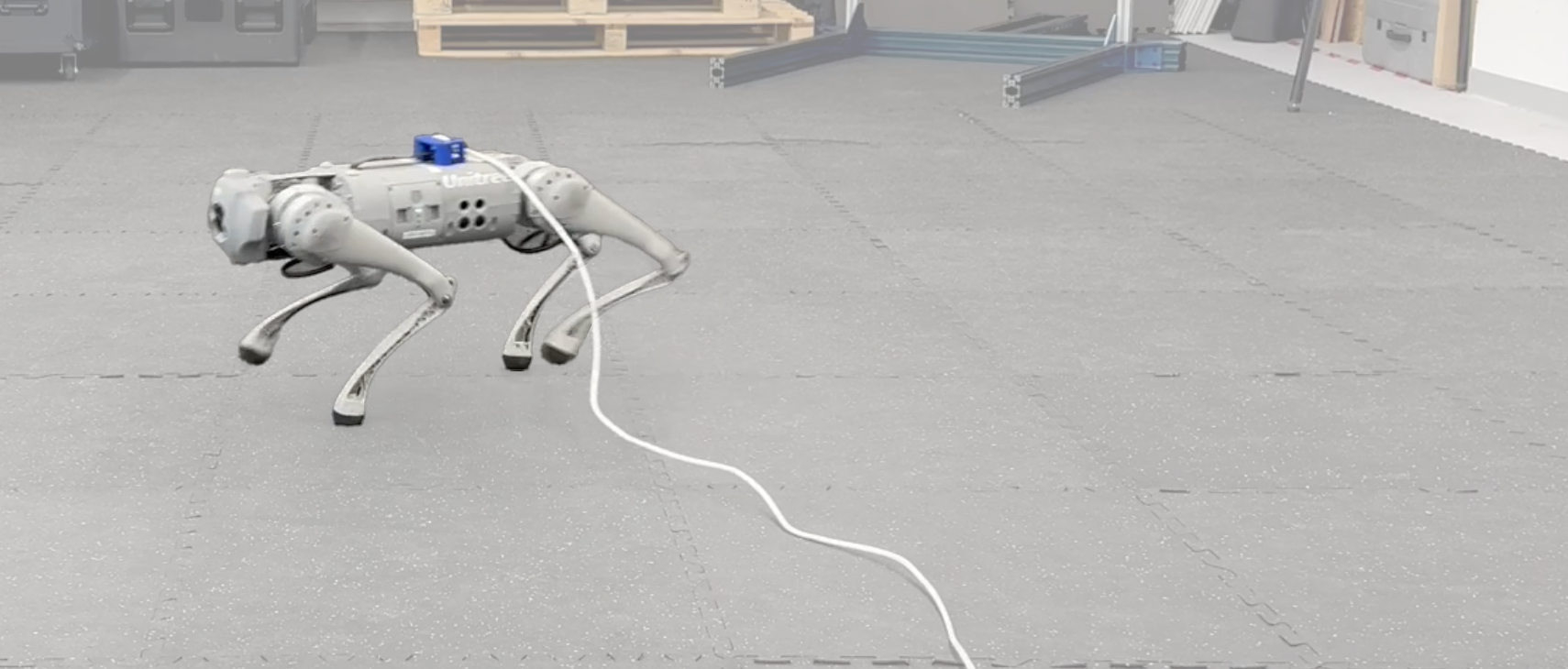}};
            \node[anchor=south west, font=\large\bfseries] at (image6.south west) {t = 2.5[sec]};
        \end{tikzpicture} &
        \begin{tikzpicture}
            \node[anchor=south west, inner sep=0] (image8) at (0, 0) 
                {\includegraphics[width=0.233\textwidth, trim={0cm 0cm 14cm 0cm}, clip]{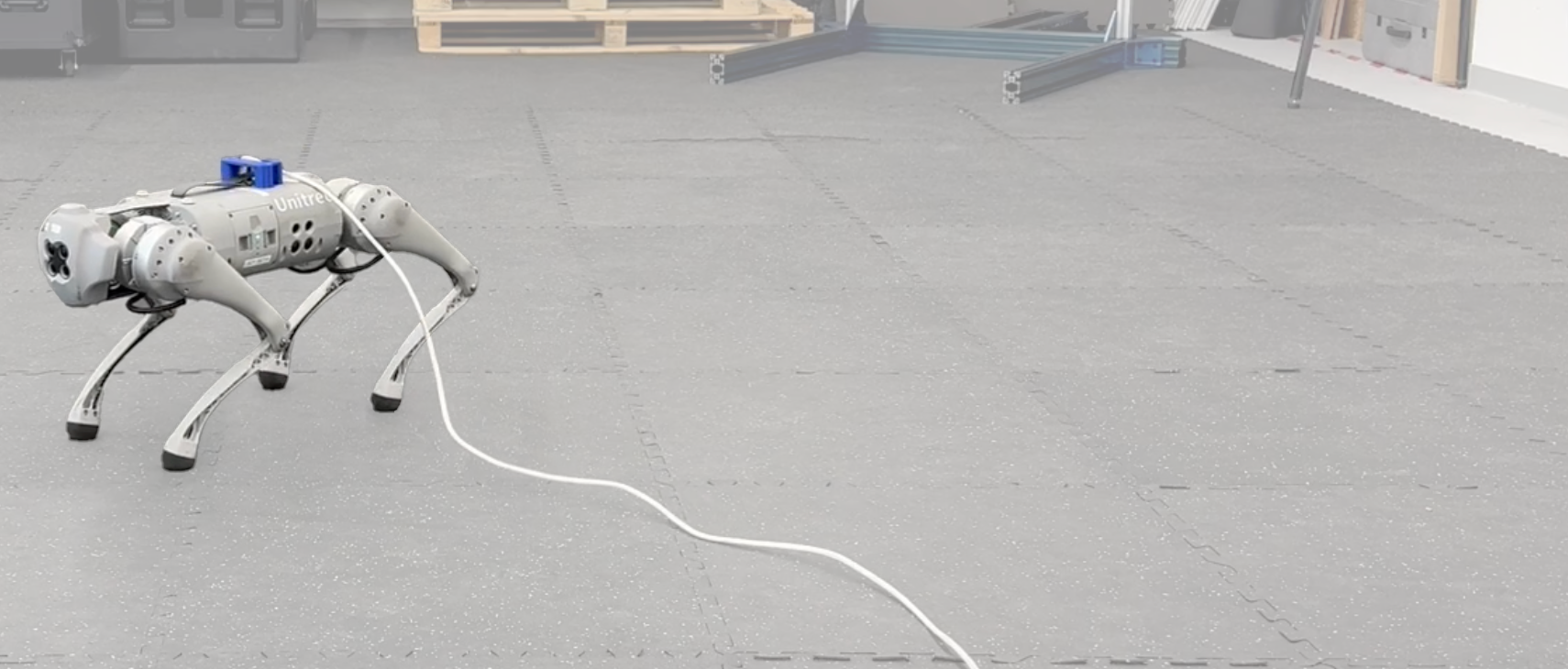}};
            \node[anchor=south west, font=\large\bfseries] at (image8.south west) {t = 3.5[sec]};
        \end{tikzpicture}
    \end{tabular}
    \caption{Example trajectories \algname{SPiDR} with RaceCar and Unitree Go1.}
    \label{fig:hardware-demo}
\end{figure}
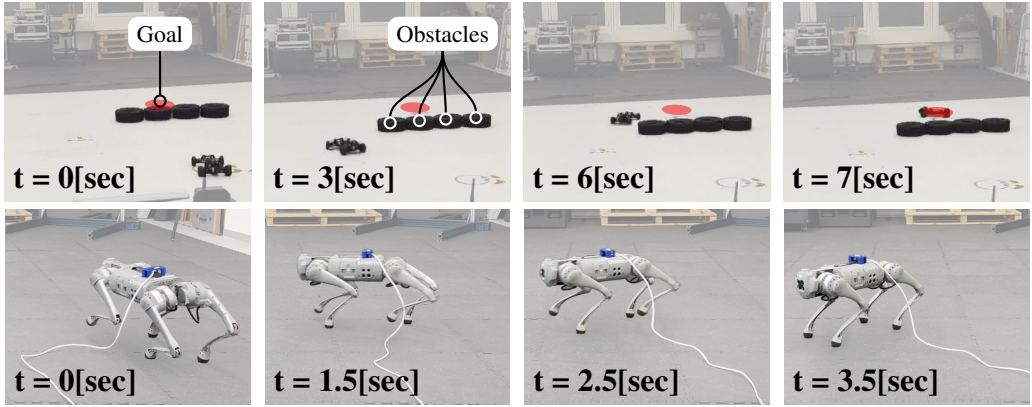

\begin{figure}
    \centering
    \includegraphics{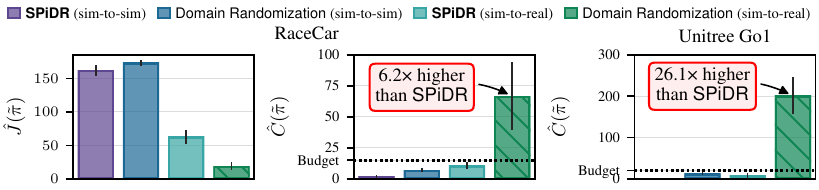}
    \caption{Performance on the race car and Unitree Go1.
    \textcolor{cool-purple}{\rule{0.6em}{0.5em}}  \algname{SPiDR} (sim-to-sim) and \textcolor{cool-teal}{\rule{0.6em}{0.5em}} \algname{SPiDR} (sim-to-real) represent evaluation \emph{in simulation} and on the \emph{real system} respectively.
    \algname{SPiDR} transfers safely, while domain randomization dramatically violates the safety constraints.
    }
    \vspace{-0.35cm}
    \label{fig:real-world-safety}
\end{figure}

\paragraph{Experiment 1: Does \algname{SPiDR} transfer safely to real systems?}
\looseness=-1We present our results in \Cref{fig:real-world-safety}. As shown, in both tasks, \algname{SPiDR} satisfies the constraint, whereas domain randomization dramatically violates it, \emph{even when the constraint is satisfied in simulation}. This result is consistent with \Cref{lem:simulation}, due to constraint underestimation in simulation. Remarkably, due to the highly dynamic behavior of the remote-control car, the performance of the objective with domain randomization underperforms compared to \algname{SPiDR}. This is because domain randomization often overshoots the target, whereas \algname{SPiDR}'s more conservative hence slower policies avoid overshooting. 
\begin{wrapfigure}[15]{r}{0.35\textwidth}
\vspace{-0.5\baselineskip}
  \begin{center}
        \includegraphics{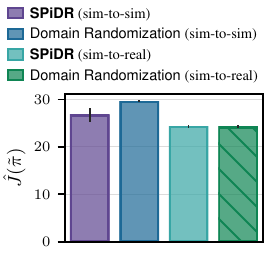}
  \end{center}
  \caption{Performance is maintained on the Unitree Go1.}
  \label{fig:go1-rewards}
\end{wrapfigure}
To evaluate the performance on the Unitree Go1 robot, we measure the rewards on the real system are report the performance in \Cref{fig:go1-rewards}. As demonstrated in \Cref{fig:go1-rewards}, both domain randomization and \algname{SPiDR} exhibit comparable decrease in performance relative to their simulated counterparts. These results demonstrate that \algname{SPiDR} transfers effectively to the real-world system without being overly conservative, successfully solving the task. We additionally provide five trajectories for each policy and algorithm, resulting in a total of 75 video demonstrations, provided in \href{https://github.com/yardenas/safe-learning}{the following link}. These recordings indicate that using \algname{SPiDR} does not lead to a noticeable degradation in locomotion performance. In comparison, \algname{RAMU}, trained with its default hyperparameters, succeeds in following the commands in $12 \pm 4.8\%$ of the trials, falling in the rest, while \algname{SPiDR} completes all trials successfully without falling. See \Cref{sec:go1-locomotion} for more details.
These results suggest that \algname{SPiDR} works well across different real-world robotic tasks, satisfying the constraints while maintaining strong performance.

\begin{wrapfigure}[14]{r}{0.35\textwidth}
\vspace{0.5\baselineskip}
  \begin{center}
        \includegraphics{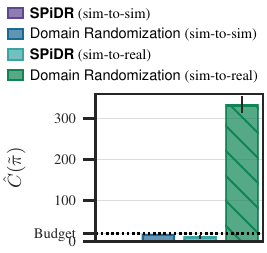}
  \end{center}
  \caption{Safe transfer to a real Unitree Go1 with PPO.}
  \label{fig:ppo-go1}
\end{wrapfigure}

\paragraph{Experiment 2: How versatile is \algname{SPiDR}?}
We replace SAC with PPO \citep{schulman2017proximal} and a primal-dual optimizer as a CMDP solver, repeating our experiment on the Unitree Go1 robot.
We present our results in \Cref{fig:ppo-go1}, demonstrating that \algname{SPiDR} satisfies the constraints on the real system, while domain randomization with PPO satisfies the constraint \emph{only in simulation}.
We provide additional sim-to-sim experiments with PPO in \Cref{sec:ppo}. Throughout this work, we use three different CMDP solvers, with two different policy search methods, demonstrating competitive performance while remaining safe, including on real hardware. These results highlight the versatility and broad applicability of \algname{SPiDR} across different CMDP solvers and RL algorithms.

\subsection{Evaluation in the Sim-to-Sim Sandbox}
\label{sec:sim-to-sim}
Next, we demonstrate that \algname{SPiDR} consistently achieves strong task performance while adhering to safety constraints in a series of sim-to-sim experiments. This set of experiments serves as a controlled testbed for systematically ablating \algname{SPiDR}. We provide additional ablations on the robustness to choice of $\lambda$ empirical intuition about $\upsilon(\cdot, \cdot)$ in \Cref{sec:additional-ablations}.
Full learning curves and further details on the experimental setup can be found in \Cref{sec:learning-curves,sec:safety-gym,sec:rwrl,sec:racecar}.

\begin{figure}
    \centering
    \includegraphics{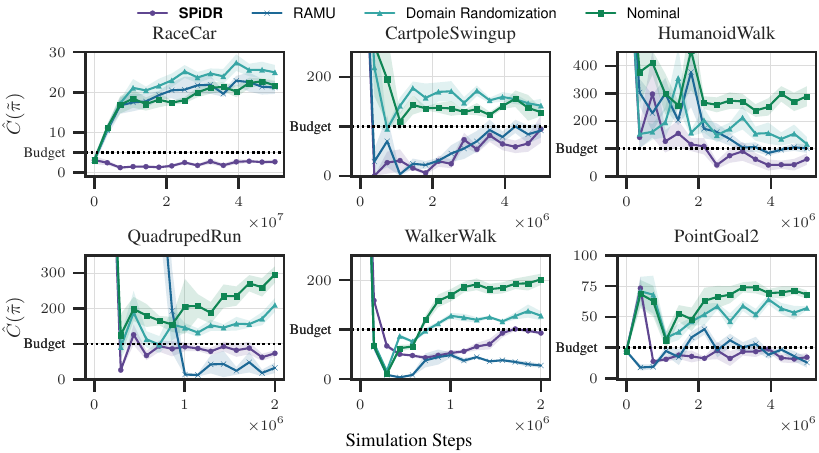}
    \caption{Costs over training iterations, evaluated on the simulated test environments. \algname{SPiDR} consistently satisfies the constraints across all tasks.}
    \vspace{-0.5cm}
    \label{fig:simulated-costs}
\end{figure}

\paragraph{Experiment 3: Can \algname{SPiDR} satisfy the constraints under sim-to-sim gap?}
\looseness=-1We demonstrate our results in \Cref{fig:simulated-costs}, reporting the learning curves of constraint across tasks.
\Cref{fig:simulated-costs} demonstrates that \algname{SPiDR} satisfies the constraints on all six tasks. This is in contrast to \algname{Nominal} and \algname{Domain Randomization}, which fail to satisfy the constraints across all tasks, as expected. In the RaceCar task, where the sim-to-sim gap primarily arises due to the training simulator failing to capture all the relevant phenomena present in the test environment, \algname{SPiDR} is the only algorithm that satisfies the constraint.

\begin{figure}
    \centering
    \includegraphics{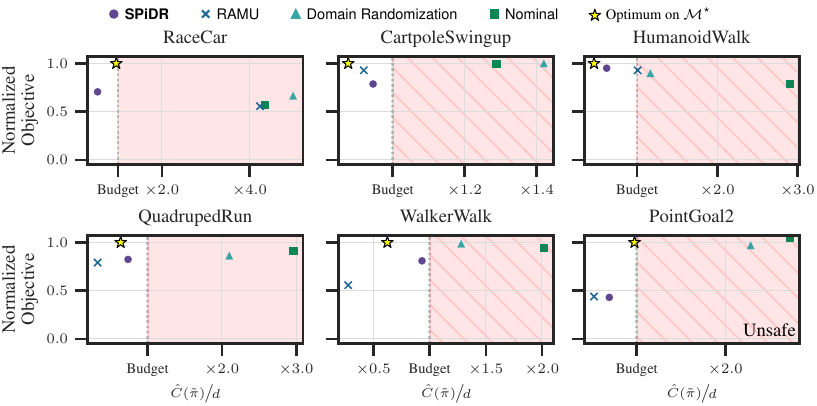}
    \caption{Normalized objective and constraint performance after training. The constraint is normalized relative to the budget, while the objective is normalized against the optimal performance on $\cM^\star$. Shaded red area represents unsafe policies. \algname{SPiDR} consistently satisfies the constraints while achieving competitive performance on different locomotion and navigation tasks.}
    \vspace{-0.3cm}
    \label{fig:simulated}
\end{figure}

\paragraph{Experiment 4: What is the tradeoff between safety and performance?}
We study the tradeoff between constraint satisfaction and objective performance, showing that \algname{SPiDR} does not suffer significant performance decrease.
We present our results in \Cref{fig:simulated}. 
As shown, \algname{SPiDR} maintains balance of high performance while satisfying the constraints across all tasks. \algname{SPiDR} and \algname{RAMU} demonstrate comparable performance on the RWRL environments. On the other hand, on the RaceCar task, which involves more realistic and challenging model discrepancies, \algname{SPiDR} significantly outperforms \algname{RAMU}. Notably, in environments like HumanoidWalk, QuadrupedRun, WalkerWalk and RaceCar, \algname{SPiDR} not only meets the safety constraints but also achieves competitive performance with respect to the optimal performance on these tasks. These results illustrate that performance is not significantly compromised, even in the presence of large model mismatch.

\begin{figure}
    \centering
    \includegraphics{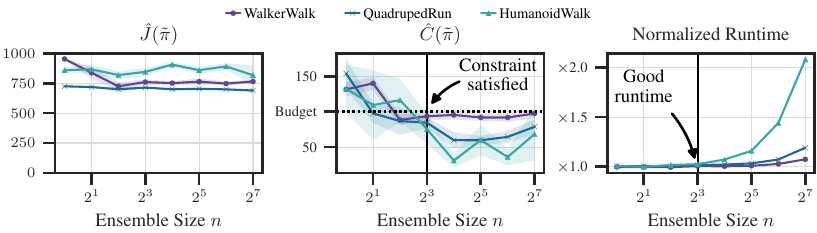}
    \caption{Runtime overhead of \algname{SPiDR} is marginal even for $n \ge 8$. For $n \ge 8$ safety is maintained across environments.}
    \vspace{-0.6cm}
    \label{fig:scaling}
\end{figure}

\paragraph{Experiment 5: How does \algname{SPiDR} scale?}
\looseness=-1We analyze how the performance of \algname{SPiDR} is affected by the ensemble size $n$ in terms of training wall-clock time and performance. We test \algname{SPiDR} on WalkerWalk, QuadrupedRun and HumanoidWalk while varying $n \in \{1, 2, 4, 8, 16, 32, 64, 128\}$. In \Cref{fig:scaling} we report the objective, constraint and the relative training runtime compared to standard domain randomization. 
As shown, for $n \ge 8$, the constraint remains within budget in all environments, while increasing $n$ has little effect on the objective. Furthermore, on all three environment the relative runtime for $n \le 32$ is only marginally longer compared to standard training with domain randomization. All experiments are run using MuJoCo XLA \citep{freeman2021brax}, enabling us to train SAC for 5M and PPO for 200M environment steps in under an hour on a single NVIDIA RTX 4090 GPU.

\begin{figure}[h]
    \centering
    \includegraphics{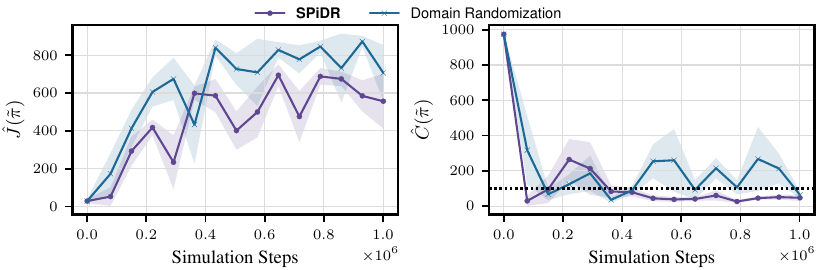}
    \caption{Learning curves of vision-based CartpoleSwingup task. \algname{SPiDR} satisfies the constraints under dynamics mismatch even when the policy observes grayscale images of the cartpole.}
    \label{fig:cartpole-vision}
\end{figure}

\paragraph{Does \algname{SPiDR} scale to vision control tasks?}
We investigate whether \algname{SPiDR} scales to partially observable vision-based settings, where the policy operates directly on rendered images. To evaluate this, we implement an asymmetric teacher–student \cite{joonho2020} setup: the policy receives only image observations, while the cost is penalized using privileged state information available in simulation, following \Cref{eq:variance-estimator}. Specifically, we combine DrQ \citep{yarats2021image} with CRPO~\citep{xu2021crpo} as the penalizer and apply \algname{SPiDR} on the CartpoleSwingup task. This setup follows our previous sim-to-sim experiment on the CartpoleSwingup task, only differing in the observation given to the agent. The policy trains on a sequence of 3 stacked grayscale 64$\times$64 pixels images, while the cost penalty uses the true system states (not shown to the agent). We report our results on the evaluation environment in \Cref{fig:cartpole-vision}.
These results show that even in a partially observable setup, \algname{SPiDR} satisfies the constraint by leveraging privileged information during training. Training the vision-based policy for 1M simulation steps with \algname{SPiDR} takes roughly 33 minutes on an RTX 4090 GPU, compared to roughly 30 minutes without \algname{SPiDR}. The main computational bottleneck lies in computing critic gradients, while simulation efficiency comes from Madrona-MJX \citep{rosenzweig24madronarenderer}, which renders rollouts from 128+ environments in parallel. This experiment suggests that \algname{SPiDR} scales effectively to vision-based control tasks.

\vspace{0.5cm}
\section{Conclusions}
\label{sec:conclusions}
In this work we address safe sim-to-real transfer, a key challenge limiting the broader adoption of RL in real-world applications. We theoretically link constraint violations upon deployment to uncertainty about the sim-to-real gap and propose a simple and provably safe algorithm that penalizes the cost function using an estimated gap under domain randomization. Our proposed method -- \algname{SPiDR} -- is theoretically sound, easy to implement and can be readily combined with popular safe RL algorithms. We empirically show that \algname{SPiDR} consistently achieves strong performance while maintaining safety across different tasks spanning three different simulated safe RL benchmarks. Moreover, \algname{SPiDR} solves two real-world robotic tasks as it ensures safe transfer, proving its practical applicability to the full-scale problem it aims to address. Although we focus our experiments on robotic tasks, due to the simplicity of \algname{SPiDR}, we believe that future work can extend it to other safety-critical domains. Finally, since our method focuses on the ``zero-shot'' transfer setting, where access to real-world data is restricted, proposing hybrid approaches that combine simulation-trained policies with safe online exploration techniques is an important direction for future work.

\newpage

\begin{ack}
We would like to thank Taerim Yoon, Yonathan Efroni, Kishan Panaganti, Chenhao Li and James Queeney for insightful discussions during the development of this project. We thank the anonymous reviewers for their valuable comments and suggestions.
This project has received funding from a grant of the Hasler foundation~(grant no. 21039) and the European Research Council (ERC) under the European Union's Horizon 2020 research and innovation programme (grant agreement No. 866480). Adam Wierman is supported by NSF through CNS-2146814, CPS-2136197, CNS-2106403, NGSDI2105648, IIS-2336236, and by the Resnick Sustainability Institute at Caltech.

\end{ack}

\clearpage
\medskip

\bibliography{references}
\bibliographystyle{plainnat}

\clearpage



\newpage

\newboolean{showsection}
\setboolean{showsection}{false}
\ifthenelse{\boolean{showsection}}{%
\section*{NeurIPS Paper Checklist}

\begin{enumerate}

\item {\bf Claims}
    \item[] Question: Do the main claims made in the abstract and introduction accurately reflect the paper's contributions and scope?
    \item[] Answer: \answerYes{} 
    \item[] Justification: In the abstract, we claim to have developed a novel algorithm with theoretical safety guarantees and a comprehensive empirical study of it. Our theory can be found in \Cref{sec:method}, with proofs in \Cref{sec:proofs,sec:heuristic}. In addition, we claim to have conducted experiments on several simulated benchmarks and on two real-world robotic platforms. These results can be found in \Cref{sec:experiments} and \Cref{sec:ppo,sec:additional-ablations,sec:learning-curves,sec:go1-locomotion,sec:racecar-real}.

\item {\bf Limitations}
    \item[] Question: Does the paper discuss the limitations of the work performed by the authors?
    \item[] Answer: \answerYes{} 
    \item[] Justification: We explicitly detail the limitations of our work, including possible venues for how they can be addressed by future work in \Cref{sec:conclusions}.

\item {\bf Theory assumptions and proofs}
    \item[] Question: For each theoretical result, does the paper provide the full set of assumptions and a complete (and correct) proof?
    \item[] Answer: \answerYes{} 
    \item[] Justification: Our main theoretical results are presented in \Cref{sec:method}. Formal assumptions required for these results are stated explicitly in \Cref{sec:proofs,sec:heuristic}. The proofs for our main theoretical result (\Cref{lem:simulation,thm:safety-performance}) are presented in \Cref{sec:proofs}. Additionally, we further justify our practical implementation in \Cref{sec:heuristic}.

    \item {\bf Experimental result reproducibility}
    \item[] Question: Does the paper fully disclose all the information needed to reproduce the main experimental results of the paper to the extent that it affects the main claims and/or conclusions of the paper (regardless of whether the code and data are provided or not)?
    \item[] Answer: \answerYes{} 
    \item[] Justification: We provide the open-source code used to run our experiments as well as exhaustive details on our experimental setup in \Cref{sec:racecar,sec:racecar-real,sec:rwrl,sec:safety-gym,sec:go1-locomotion}. The physical remote control car experiments are unfortunately not trivial to reproduce, since they rely on the licensed software for the motion capture system. However, the training code for this environment is open-sourced and the experimental setup itself is fairly simple. For the Unitree Go1 robot, aside from providing the code used for the experiments, we provide the policies we deployed on the real platform in \url{https://anonymous.4open.science/r/safe-sim2real-1EAC}.

\item {\bf Open access to data and code}
    \item[] Question: Does the paper provide open access to the data and code, with sufficient instructions to faithfully reproduce the main experimental results, as described in supplemental material?
    \item[] Answer: \answerYes{}
    \item[] Justification: A URL with our anonymized open-source code is provided in \url{https://anonymous.4open.science/r/safe-sim2real-1EAC}. The link includes specific installation instructions. Furthermore, for the Unitree Go1 experiments, we provide the policies used on hardware in onnx format, essentially allowing anyone with this robot to evaluate the policies used in our experiments.

\item {\bf Experimental setting/details}
    \item[] Question: Does the paper specify all the training and test details (e.g., data splits, hyperparameters, how they were chosen, type of optimizer, etc.) necessary to understand the results?
    \item[] Answer: \answerYes{}
    \item[] Justification: All training details for the experiments are provided in \Cref{sec:racecar,sec:racecar-real,sec:rwrl,sec:safety-gym,sec:go1-locomotion}.

\item {\bf Experiment statistical significance}
    \item[] Question: Does the paper report error bars suitably and correctly defined or other appropriate information about the statistical significance of the experiments?
    \item[] Answer: \answerYes{} 
    \item[] Justification: We explicitly mention in \Cref{sec:experiments} that we report the mean and standard error in all of our reported results. Our hardware experiments are evaluated across several seeds and trials for each experiment.

\item {\bf Experiments compute resources}
    \item[] Question: For each experiment, does the paper provide sufficient information on the computer resources (type of compute workers, memory, time of execution) needed to reproduce the experiments?
    \item[] Answer: \answerYes{} 
    \item[] Justification: We explicitly mention in \Cref{sec:experiments} the compute used in all of our experiments.
    
\item {\bf Code of ethics}
    \item[] Question: Does the research conducted in the paper conform, in every respect, with the NeurIPS Code of Ethics \url{https://neurips.cc/public/EthicsGuidelines}?
    \item[] Answer: \answerYes{} 
    \item[] Justification: The research conducted in this work fully follows the ``NeurIPS Code of Ethics''.

\item {\bf Broader impacts}
    \item[] Question: Does the paper discuss both potential positive societal impacts and negative societal impacts of the work performed?
    \item[] Answer: \answerYes{} 
    \item[] Justification: Our algorithm \algname{SPiDR}, aims to make RL agents safe and trustworthy, potentially unlocking many domains in which RL can be deployed reliably \citep[e.g. in healthcare,][]{pmlr-v126-fox20a,s20185058} and improve our lives. While malicious usage cannot be strictly excluded, we believe that if \algname{SPiDR} is used \emph{bona fide}, its overall societal impact will be net positive. We discuss this in furthre details in \Cref{sec:introduction}.
    
\item {\bf Safeguards}
    \item[] Question: Does the paper describe safeguards that have been put in place for responsible release of data or models that have a high risk for misuse (e.g., pretrained language models, image generators, or scraped datasets)?
    \item[] Answer: \answerNA{} 
    \item[] Justification: Our paper does not pose such risks.

\item {\bf Licenses for existing assets}
    \item[] Question: Are the creators or original owners of assets (e.g., code, data, models), used in the paper, properly credited and are the license and terms of use explicitly mentioned and properly respected?
    \item[] Answer: \answerYes{} 
    \item[] Justification: We have explicitly referenced all works upon which this research is based; for example, we have credited the authors of MuJoCo appropriately throughout the paper.

\item {\bf New assets}
    \item[] Question: Are new assets introduced in the paper well documented and is the documentation provided alongside the assets?
    \item[] Answer: \answerYes{} 
    \item[] Justification: Our open-source code includes a README file with specific instructions for installation and usage.

\item {\bf Crowdsourcing and research with human subjects}
    \item[] Question: For crowdsourcing experiments and research with human subjects, does the paper include the full text of instructions given to participants and screenshots, if applicable, as well as details about compensation (if any)? 
    \item[] Answer: \answerNA{} 
    \item[] Justification: Our work does not involve crowdsourcing nor research with human subjects.

\item {\bf Institutional review board (IRB) approvals or equivalent for research with human subjects}
    \item[] Question: Does the paper describe potential risks incurred by study participants, whether such risks were disclosed to the subjects, and whether Institutional Review Board (IRB) approvals (or an equivalent approval/review based on the requirements of your country or institution) were obtained?
    \item[] Answer: \answerNA{} 
    \item[] Justification:  Our work does not involve crowdsourcing nor research with human subjects.

\item {\bf Declaration of LLM usage}
    \item[] Question: Does the paper describe the usage of LLMs if it is an important, original, or non-standard component of the core methods in this research? Note that if the LLM is used only for writing, editing, or formatting purposes and does not impact the core methodology, scientific rigorousness, or originality of the research, declaration is not required.
    \item[] Answer: \answerNA{} 
    \item[] Justification: We used LLMs only for language editing purposes.

\end{enumerate}

\newpage
}{}

\appendix
\section*{Appendix}

\section{Failure modes of Domain Randomization}
\label{sec:dr-failure-modes}
The following example illustrates how domain randomization can fail to ensure constraint satisfaction in practice, even if all the simulator environments are arbitrarily close to the real environment.
\begin{example}
\label{example:worst-case}
    Suppose $\cS\ge 3$, $\cA\ge 2$, then for all $0<\varepsilon\le 1/4$, there exists a set of simulated CMDPs $\{\cM_i=(\cS, \cA, \hat{p}_{\xi_i}, r, c, \gamma, \rho)\}_{i=1}^N$, an unknown real CMDP $\cM^\star=(\cS, \cA, p^\star, r, c, \gamma, \rho)$ satisfying
    \begin{equation*}
        D_W(p^\star,\hat{p}_{\xi_i})(s,a)\le \varepsilon,\ \forall i\in\{1,\ldots,N\}, (s,a)\in \cS\times\cA.
    \end{equation*}
    with a fixed budget $d>0$ for all CMDPs, such that the domain randomization policy $\pi_{DR}$ returned by \Cref{eq:DR-average} with any $\mu$, will be unsafe in the real environment,
    \begin{equation*}
        C_{p^\star}(\pi_{DR}) > d.
    \end{equation*}
\end{example}
\begin{proof}
    Consider the CMDPs given in \Cref{fig:worst-case}, where there are three states, the initial state $s_0$, two absorbing states $s_1$ and $s_2$, two actions $a_1$ and $a_2$. The state space is in $\mathbb{R}$, and we assume $s_0=0$, $s_1=1$ and $s_2=2$. The reward and cost function are given by:
    \begin{align*}
        r(s_0,a)&=0,\ c(s_0,a)=0,\ \forall a \in \cA;\\
        r(s_1,a)&=1,\ c(s_1,a)=1,\ \forall a \in \cA;\\
        r(s_2,a)&=0,\ c(s_2,a)=0,\ \forall a \in \cA;
    \end{align*}
    In the $i$-th simulated environment, at the initial state $s_0$, the transition probability is given by:
    \begin{equation*}
    \left\{
    \begin{aligned}
      & \hat{p}_{\xi_i}(s_0 \mid  s_0, a) = 0,\,\forall\,a\in \cA, \\
      & \hat{p}_{\xi_i}(s_1 \mid  s_0, a) = \frac{1}{2}+\varepsilon\indic{a=a_1},\,\forall\,a\in \cA, \\
      & \hat{p}_{\xi_i}(s_2 \mid  s_0, a) = \frac{1}{2} - \varepsilon\indic{a=a_1}),\,\forall\,a\in\cA,
    \end{aligned}
    \right.
    \end{equation*}
    In the real environment, at the initial state $s_0$, the transition probability is given by:
    \begin{equation*}
    \left\{
    \begin{aligned}
      & p^\star(s_0 \mid  s_0, a) = 0,\,\forall\,a\in \cA, \\
      & p^\star(s_1 \mid  s_0, a) = \frac{1}{2}+\varepsilon(1+\indic{a=a_1}),\,\forall\,a\in\cA, \\
      & p^\star(s_2 \mid  s_0, a) = \frac{1}{2} - \varepsilon(1+\indic{a=a_1}),\,\forall\,a\in\cA,
    \end{aligned}
    \right.
    \end{equation*}
    For all these environments, at the absorbing states $s_1$ and $s_2$, we have $p(s_2\mid s_2,a)=p(s_1\mid s_1,a)=1$ for $\forall a\in\cA$. Then the maximum Wasserstein distance  can be bounded by epsilon, i.e. $D_W(p^\star,\hat{p}_{\xi_i})(s,a)\le \varepsilon, \forall(s,a)\in \cS\times\cA$ for every simulated environment. 
    Take the budget to be $d=\frac{\gamma}{1-\gamma}(\frac{1}{2}+\varepsilon)$.
    
    To achieve the highest cumulative reward, the optimal domain randomization policies returned by \Cref{eq:DR-average} is supported only on $a_1$ for any training distribution $\mu$, but the cost incurred by these policies in the real environment is $(\frac{1}{2}+2\varepsilon)\frac{\gamma}{1-\gamma}>d$, violating the constraint in the real environment.
\begin{figure}[h!]
    \centering
    \begin{tikzpicture}[font=\footnotesize]

\begin{scope}[shift={(-5cm,0)}]
    \node[rectangle, fill=blue!10, rounded corners, inner sep=5pt, draw=black, line width=1pt] 
        at (0,2.5) {\textbf{Simulated Dynamics}};

    \node[circle, draw, fill=gray!15, inner sep=1pt] (S1R) at (0, 0) {\shortstack[c]{$r=0$,\\$c=0$}};
    \node[circle, draw, inner sep=1pt] (S2R) at (-1.25, 1.25) {\shortstack[c]{$r=1$,\\$c=1$}};
    \node[circle, draw, inner sep=1pt] (S3R) at (1.25, 1.25) {\shortstack[c]{$r=0$,\\$c=0$}};

    \draw[->, thick] (S1R) -- (S2R) node[midway, below left]
        {$\frac{1}{2}+\varepsilon\mathds{1}(a=a_1)$};
    \draw[->, thick] (S1R) -- (S3R) node[midway, below right] 
        {$\frac{1}{2}-\varepsilon\mathds{1}(a=a_1)$};
    \draw[->, thick] (S2R) edge [loop left] node {$1$} ();
    \draw[->, thick] (S3R) edge [loop right] node {$1$} ();
\end{scope}

\begin{scope}[shift={(2cm,0)}]
    \node[rectangle, fill=orange!15, rounded corners, inner sep=5pt, draw=black, line width=1pt] 
        at (0,2.5) {\textbf{Real Dynamics}};

    \node[circle, draw, fill=gray!15, inner sep=1pt] (S1L) at (0, 0) {\shortstack[c]{$r=0$,\\$c=0$}};
    \node[circle, draw, inner sep=1pt] (S2L) at (-1.25, 1.25) {\shortstack[c]{$r=1$,\\$c=1$}};
    \node[circle, draw, inner sep=1pt] (S3L) at (1.25, 1.25) {\shortstack[c]{$r=0$,\\$c=0$}};

    \draw[->, thick] (S1L) -- (S2L) node[midway, below left] 
        {$\frac{1}{2}+\varepsilon(1+\indic{a=a_1})$};
    \draw[->, thick] (S1L) -- (S3L) node[midway, below right] 
        {$\frac{1}{2}-\varepsilon(1+\indic{a=a_1})$};
    \draw[->, thick] (S2L) edge [loop left] node {$1$} ();
    \draw[->, thick] (S3L) edge [loop right] node {$1$} ();
\end{scope}

\draw[->, line width=1.25pt] 
    (-3.5, 1.85) .. controls (-2.5, 2.5) and (-0.5, 2.5) .. (0.5, 1.85)
    node[midway, above, font=\itshape] {\textit{sim-to-real deployment}};

\end{tikzpicture}
    \caption{Pathological unsafe transfer. Light grey states denote the initial state $s_0$. Constraints are prone to be violated, even under a small mismatch between the simulated and the real system dynamics.}
    \label{fig:worst-case}
\end{figure}
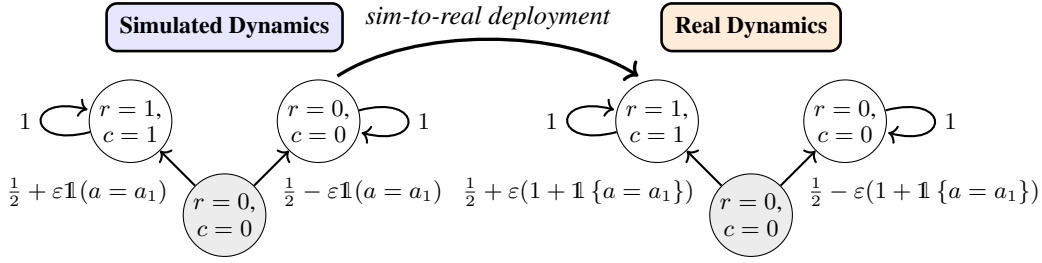
\end{proof}
We see that a policy trained via standard domain randomization in the example \Cref{example:worst-case} can still be unsafe in the real environment. The only safe policy in the real environment is $\pi(s_0)=a_2$, which cannot be learned with pure domain randomization in this example. This unsafe transfer is not only due to the averaging cost constraint, but also due to inherent mismatch between any simulated and real dynamics. This  mismatch may lead the learned policy to frequently visit regions with underestimated cost in simulation, but which incur high actual cost in the real world.

\section{Proofs}
\label{sec:proofs}
\subsection{Preliminaries}
We first present a well-established lemma, the proof of which can be found in many works, e.g. \citet[][Lemma 1.16]{agarwal_reinforcement_nodate}.
\begin{lemma}[Telescoping lemma]\label{lem:telescoping}
        Given a policy $\pi$, for different dynamics $p$ and $q$, let $g_{q,\pi,f}(s,a)\triangleq\EE_{s'\sim q(\cdot|s,a)}f^{p,\pi}(s')-\EE_{s'\sim p(\cdot|s,a)}f^{p,\pi}(s')$, where $f$ is used to overload the notation for the constraint function $V_c$. Then, we have
        \begin{align*}
            C_{q}(\pi )-C_{p}(\pi )&=\frac{\gamma}{1-\gamma}\EE_{(s,a)\sim d_{q,\pi}}g_{q,\pi,C}(s,a).
        \end{align*}
    \end{lemma}
Starting from this point, we assume the following Lipschitz continuity condition holds throughout the remainder of this section:
\begin{assumption}[Continuity]\label{assumption:lipschitz}
For any $\pi \in \polclass$, the state cost value $V^{p,\pi}_c(s)$ and state reward value $V^{p,\pi}_r(s)$ are $L_C$- and $L_J$-Lipschitz continuous in $s\in\cS$ w.r.t. the $1$-norm respectively, over both $\realmdp$ and $\widehat{\cM}_\xi$ for all $\xi \in \Xi$.
\end{assumption}

\begin{assumption}[Finite Discrepancy]\label{assumption:estimator}
We assume that the worst-case model discrepancy between the real and all simulated environments for all $(s,a) \in \cS \times \cA$ is finite, i.e.,
    \begin{equation*}
        \max_{\xi\in\Xi}D_W(\hat{p}_\xi,p^\star)(s,a) < \infty,\ \forall (s,a)\in\cS\times\cA.
    \end{equation*}
\end{assumption}
\subsection{Proof of \Cref{lem:simulation}}
\begin{proof}
    Given a policy $\pi$, for dynamics $p^\star$ and $ \hat{p}_\xi$, let $g_{ \hat{p}_\xi,\pi,C}(s,a)\triangleq\EE_{s'\sim  \hat{p}_\xi(\cdot|s,a)}V^{p^\star,\pi}_c(s')-\EE_{s'\sim p^\star(\cdot|s,a)}V^{p^\star,\pi}_c(s')$. Let $||\cdot||_{Lip}$ denote the Lipschitz constant of a function. Under the assumption that the state cost function is Lipschitz, by the Kantorovich-Rubinstein representation \citep[Section 11.8]{Dudley_2002} for $L_1$ Wasserstein distance, we have:
    \begin{align}
        g_{ \hat{p}_\xi,\pi,C}&\le L_C\sup_{||f||_{Lip}\le 1}|\EE_{s'\sim p^\star(\cdot|s,a)}f(s')-\EE_{s'\sim  \hat{p}_\xi(\cdot|s,a)}f(s')|\nonumber\\
        &\le L_C\cdot D_W(\hat{p}_\xi, p^\star)(s,a).\label{eq:gC-to-dW}
    \end{align}
    By \Cref{lem:telescoping}, we have
    \begin{align}
        \left|C_{p^\star}(\pi )-\EE_{\xi\sim \mu}C_{ \hat{p}_\xi}(\pi )\right|&\leq\left|\EE_{\xi\sim \mu}[C_{p^\star}(\pi)-C_{ \hat{p}_\xi}(\pi)]\right| \nonumber \\
        &=\left|\frac{\gamma}{1-\gamma}\EE_{\xi\sim \mu}\bigg[\EE_{(s,a)\sim d_{ \hat{p}_\xi,\pi}}[g_{ \hat{p}_\xi,\pi,C}(s,a)]\bigg]\tag{\Cref{lem:telescoping}}\right|\\
        &\le\frac{\gamma L_C}{(1-\gamma)}\EE_{\xi\sim \mu}\bigg[\EE_{(s,a)\sim d_{ \hat{p}_\xi,\pi}}[D_W(\hat{p}_\xi, p^\star)(s,a)]\bigg] \tag{\Cref{eq:gC-to-dW}}.
    \end{align}
    This ends the proof of \Cref{lem:simulation}.
\end{proof}

\subsection{Proof of \Cref{thm:safety-performance}}

\begin{proof}
    We first prove that $\EE_{\xi\sim\mu}[\tilde{C}_{ \hat{p}_\xi}(\pi )]\ge C_{p^\star}(\pi )$ for any policy $\pi\in\polclass$. By definition of the penalized cost $\tilde{C}_{ \hat{p}_\xi}(\pi)$ and using \Cref{lem:simulation}, we have
    \begin{align}\label{eq:geq-z}
        &\EE_{\xi\sim\mu}[\tilde{C}_{ \hat{p}_\xi}(\pi)] - C_{p^\star}(\pi) \nonumber\\
        &=\EE_{\xi\sim \mu}[C_{ \hat{p}_\xi}(\pi )] - C_{p^\star}(\pi ) \nonumber \\
        & \quad+\EE_{\xi\sim \mu}\bigg[\frac{\gamma L_C}{1-\gamma}\EE_{(s,a)\sim d_{ \hat{p}_\xi,\pi}}[\max_{\xi'\in\Xi}D_W(\hat{p}_{\xi'}, p^\star)(s,a)]\bigg] \nonumber\\
        &\ge - \frac{\gamma L_C}{(1-\gamma)}\EE_{\xi\sim \mu}\bigg[\EE_{(s,a)\sim d_{ \hat{p}_\xi,\pi}}[D_W(\hat{p}_\xi, p^\star)(s,a)]\bigg] \nonumber\\
        & \quad + \frac{\gamma L_C}{(1-\gamma)}\EE_{\xi\sim \mu}\bigg[\EE_{(s,a)\sim d_{ \hat{p}_\xi,\pi}}[\max_{\xi'\in\Xi}D_W(\hat{p}_{\xi'}, p^\star)(s,a)]\bigg] \tag{\Cref{lem:simulation}}\\
        &\ge 0.
    \end{align}
    Where in the last inequality we used that $D_W(\hat{p}_\xi, p^\star)(s,a) \leq \max_{\xi'\in\Xi}D_W(\hat{p}_{\xi'}, p^\star)(s,a)$.
    Next, since $\EE_{\xi\sim\mu}[\tilde{C}_{\hat{p}_\xi}(\tilde{\pi})]\le d$, and with \Cref{eq:geq-z} directly gives us:
    \begin{equation*}
        C_{p^\star}(\tilde{\pi})\le \EE_{\xi\sim\mu}[\tilde{C}_{\hat{p}_\xi}(\tilde{\pi})]\le d.
    \end{equation*}
    This ends the proof of \Cref{eq:safety}.
\end{proof}

\section{Designing $\boldsymbol{\upsilon(s, a)}$}
\label{sec:heuristic}
\looseness=-1In this section we provide additional theoretical intuition for the variance estimator introduced in \Cref{eq:variance-estimator}. Next, we show that, under that under the assumptions stated below, $\upsilon(\cdot, \cdot)$ upper-bounds the true sim-to-real discrepancy, measured by the $L_1$-Wasserstein distance, with high probability, using  data collected in simulation. Importantly, when $\upsilon(\cdot, \cdot)$ is indeed an upper bound for the true model discrepancy, \algname{SPiDR} is \emph{provably} guaranteed to be safe, as described above in \Cref{sec:proofs}.
\subsection{Modeling Assumptions}
Let $d(\cdot,\cdot)$ be the Euclidean $2$-norm on~$\cS$, and let
$p_\mu \triangleq \mathbb{E}_{\xi\sim\mu}[p_\xi]$ denote the
\emph{domain-randomization kernel}.
\begin{definition}[$L_1$-Wasserstein distance \citep{clark}]\label{definition:wasserstein}
    Given a $\sigma$-algebra $\cF$,  for any two probability measures $P_1, P_2\in\cM(\cS,\cF)$, the $L_1$-Wasserstein distance between them is defined as:
    \begin{equation}\label{eq:def-Wasserstein}
        D_W(P_1,P_2)\triangleq\inf_{\gamma\in\Gamma(P_1,P_2)}\int_{\cS\times\cS}d(s_1,s_2)\gamma(ds_1\times ds_2),
    \end{equation}
where $\Gamma(P_1,P_2)\triangleq\{\gamma\in\cM(\cS\times\cS,\cF\times\cF):\gamma(A\times \cS)=P_1(A),\gamma(\cS\times A)=P_2(A),\forall A\in\cF\}$ is the set of all couplings of $P_1$ and $P_2$, and $d(\cdot,\cdot)$ is a $\cF\times\cF$-measurable metric defined on $\Xi$. In our setting, we assume this metric is given by a 2-norm $||\cdot||_2$. 
\end{definition}
\begin{assumption}[Mild sim-to-real gap]\label{assumption:sim2real-degree}
    There exists a constant $\varepsilon>0$, such that for any $(s,a)\in\cS\times \cA$, $\mathrm{KL}(p^\star,p_\mu)\le \varepsilon$ and $\max_{\xi\in\Xi} \mathrm{KL}(p_\xi, p_\mu)\le \varepsilon$.
\end{assumption}This assumption states that the real dynamics are close to the 
domain-randomization kernel and that the domain-randomization
support~$\Xi$ is not overly large.  
In practice, practitioners design~$\Xi$ using the domain knowledge of the
real system; the uniform KL radius formalizes this intuition. Additionally, assuming that $\mathrm{KL}(p^\star,p_\mu)\le \varepsilon$ ensures that the simulator provides a reasonably accurate approximation of the real system. While the Wasserstein distance $D_W$ serves as the primary metric for model mismatch in our analysis, we use the KL divergence as it yields tighter concentration bounds, which we use below.
\begin{assumption}[Bounded state space]\label{assumption:bounded-space}
    The state space $\cS$ is compact with diameter $d_s$.
\end{assumption}
This is a fairly mild assumption in many physical problems, common for example in robotics.
We next assume the sample average distance to the average next state in \Cref{alg:spidr} is uniformly lower bounded.
\begin{assumption}[Non-degenerate variance]\label{assumption:regularization}
    Let $s_i$ be $n$ i.i.d. draws from $p_{\mu}(\cdot\mid s,a)$ and $\overline{s}_{s,a}\triangleq \frac{1}{n}\sum_{i=1}^n s_i$ their empirical average. Furthermore, let $\hat{p}_{\mu}(s'\mid s,a)\triangleq\frac{1}{n}\sum_{i=1}^n \mathds{1}[s'=s_i]$ denote the empirical measure associated with the sample set. We assume the expected distance between $s_i$ and $\overline{s}_{s,a}$ is uniformly lower bounded by a constant $c_3$, i.e.,
    \begin{equation*}
        \EE_{s'\sim \hat{p}_\mu(\cdot\mid s,a)}[d(s',\overline{s}_{s,a})]\ge c_3,\ \forall(s,a)\in\cS\times\cA.
    \end{equation*}
\end{assumption}
If the next-state distribution collapses to a point
($c_3\!\approx\!0$) the empirical variance trivially underestimates any
model mismatch. Requiring a minimal spread rules out this
degenerate case. In addition, many real systems are not fully deterministic and exhibit process and sensor noise. Hence a uniform lower bound $c_3$ is both mild and practically satisfiable.

\subsection{Auxiliary Lemmas}
The following lemmas are used in our analysis below.
\begin{lemma}[Bernstein transportation \citep{talebi2018variance}]\label{lem:trans}
Let $ p, q \in \Sigma_S $, where $\Sigma_S$ denotes the probability simplex of dimension $S - 1$. For all $\alpha > 0$, for all functions $f$ defined on $\mathcal{S}$ with $0 \leq f(k) \leq b$, for all $s \in \mathcal{S}$, if $\mathrm{KL}(p, q) \leq \alpha$ then

\begin{equation*}
    |pf - qf| \leq \sqrt{2 \mathrm{Var}_q(f) \alpha} + \frac{2}{3} b \alpha,
\end{equation*}
where we use the expectation operator defined as $ pf \triangleq \mathbb{E}_{s \sim p} f(s) $ and the variance operator defined as $\mathrm{Var}_p(f) \triangleq \mathbb{E}_{s \sim p} (f(s) - \mathbb{E}_{s' \sim p} f(s'))^2 = p(f - pf)^2 $.
\end{lemma}
\begin{lemma}[Variance bound for change of measure \citep{menard_fast_2021}] \label{lem:var-pq}
Let $ p, q \in \Sigma_S $ and $f $ is a function defined on $ S $ such that $ 0 \leq f(s) \leq b $ for all $ s \in S $. If $ \mathrm{KL}(p, q) \leq \beta $ then
\begin{equation*}
    \mathrm{Var}_q(f) \leq 2 \mathrm{Var}_p(f) + 4 b^2 \beta \quad \text{and} \quad \mathrm{Var}_p(f) \leq 2 \mathrm{Var}_q(f) + 4 b^2 \beta.
\end{equation*}
\end{lemma}
\begin{lemma}[\text{\citet[Proposition 1]{jonsson2020planning}}]\label{lem:kl-divergence}
For all $ p \in \Sigma_m $ and for all $ \alpha \in [0,1] $,
\begin{equation*}
    \mathbb{P} \left( \forall n \in \mathbb{N}^+, n \mathrm{KL}(\hat{p}_n, p) \le \log\left(\frac{1}{\alpha}\right) + (m-1) \log \left( e\left(1 + \frac{n}{m-1}\right) \right) \right) \ge 1-\alpha.
\end{equation*}
\end{lemma}
\subsection{A High-probability Wasserstein bound}
We are now ready to state the theoretical guarantee for~$\upsilon(\cdot, \cdot)$,
showing that it upper-bounds the worst-case $L_1$-Wasserstein distance
between the simulated and real dynamics.
\begin{theorem}\label{thm:guarantee-of-v}
    Under Assumptions \ref{assumption:sim2real-degree}-\ref{assumption:regularization}, given a confidence level $\alpha\in(0,1)$, for any $(s,a)\in \cS\times \cA$, there exist constants $C_1,C_2,C_3,C_4>0$ that contain only $S,A,\alpha$, $d_s$, $\varepsilon$ and $\log$ factor of $n$, such that with probability at least $1-\alpha$, we have:
    \begin{equation*}
        \max_{\xi\in\Xi}D_W(p_\xi,p^\star)(s,a)\le C_1 \upsilon(s,a)+C_2\varepsilon+\sqrt{\frac{C_3}{n}}+\frac{C_4}{n}.
    \end{equation*}
\end{theorem}
The leading term $C_1\upsilon(s,a)$ captures how the \emph{local}
spread of simulated transitions bounds the worst-case
Wasserstein gap. The $\varepsilon$ term reflects irreducible model
mismatch, and the remaining terms are standard finite-sample
corrections. The key insight lies in the first term, which uses the variance in our algorithm. We note that while the design of $\upsilon(\cdot, \cdot)$ is not necessarily the tightest possible, it is easy to implement, making it more widely applicable. Our proof for \Cref{thm:guarantee-of-v} is stated below.
\begin{proof}
    We first establish an upper bound on $D_W(p_\xi,p^\star)(s,a)$ using the $2$-norm. Specifically, for each $(s,a)$, consider the coupling between $p_\xi(\cdot\mid s,a)$ and $p^\star(\cdot\mid s,a)$, defined by:
    \begin{equation*}
        \gamma(s_1,s_2)_{s,a}=p_\xi(s_1\mid s,a)\cdot p^\star(s_2\mid s,a).
    \end{equation*}
    By the definition of the $L_1$-Wasserstein distance given in \Cref{eq:def-Wasserstein}, we have:
    \begin{equation}\label{eq:dw-d2}
        D_W(p_\xi,p^\star)(s,a)\le \EE_{s_1\sim p_\xi(\cdot\mid s,a)}\bigg[\EE_{s_2\sim p^\star(\cdot\mid s,a)}[d(s_1,s_2)]\bigg].
    \end{equation}
    Recall that $\overline{s}_{s,a}\triangleq \frac{1}{n}\sum_{i=1}^n s_i$, where $s_i$ are i.i.d. drawn from $p_{\mu}(\cdot\mid s,a)$. By the triangle inequality of the $2$-norm, we have:
    \begin{equation*}
        d(s_1,s_2)\le d(s_1,\overline{s}_{s,a})+ d(s_2,\overline{s}_{s,a}).
    \end{equation*}
    Substituting this into \Cref{eq:dw-d2} yields:
    \begin{align}\label{eq:s-s_mu}
        D_W(p_\xi,p^\star)(s,a)&\le \EE_{s_1\sim p_\xi(\cdot\mid s,a)}[d(s_1,\overline{s}_{s,a})]+\EE_{s_2\sim p^\star(\cdot\mid s,a)}[d(s_2,\overline{s}_{s,a})].
    \end{align}
    Since $p_\xi$ and $p^\star$ are unknown, we approximate them using $p_\mu$ via \cref{lem:trans}. Under \Cref{assumption:sim2real-degree} and \Cref{assumption:bounded-space}, we obtain:
    \begin{align*}
        \EE_{s_1\sim p_\xi(\cdot\mid s,a)}[d(s_1,\overline{s}_{s,a})]&\le \EE_{s_1\sim p_\mu(\cdot\mid s,a)}[d(s_1,\overline{s}_{s,a})]+\sqrt{2\mathrm{Var}_{p_\mu}(d(s_1,\overline{s}_{s,a}))\varepsilon}+\frac{2}{3}d_s\cdot \varepsilon,\\
        \EE_{s_2\sim p^\star(\cdot\mid s,a)}[d(s_2,\overline{s}_{s,a})]&\le \EE_{s_2\sim p_\mu(\cdot\mid s,a)}[d(s_2,\overline{s}_{s,a})]+\sqrt{2\mathrm{Var}_{p_\mu}(d(s_2,\overline{s}_{s,a}))\varepsilon}+\frac{2}{3}d_s\cdot \varepsilon.
    \end{align*}
    Substituting into \Cref{eq:s-s_mu} gives:
    \begin{equation}\label{eq:empirical}
        D_W(p_\xi,p^\star)(s,a)\le 2\EE_{s'\sim p_\mu(\cdot\mid s,a)}[d(s',\overline{s}_{s,a})]+2\sqrt{2\mathrm{Var}_{p_\mu}(d(s',\overline{s}_{s,a}))\varepsilon}+\frac{4}{3}d_s\cdot \varepsilon
    \end{equation}
    Recall that $\hat{p}_{\mu}(s'\mid s,a)\triangleq\frac{1}{n}\sum_{i=1}^n \mathds{1}[s'=s_i]$ denote the empirical measure. By \Cref{lem:kl-divergence}, we have w.p. $1-\alpha$, for any $(s,a)$, the following inequality holds:
    \begin{equation*}
        \mathrm{KL}(\hat{p}_\mu, p_\mu)(s,a) \le \frac{1}{n} \left( \log\left(\frac{1}{\alpha}\right) + \left(S-1\right)\log\left(e\left(1+\frac{n}{S-1}\right)\right) \right)
    \end{equation*}
    Define $g(S,A,n,\alpha)\triangleq\log\left(\frac{1}{\alpha}\right)+(S-1)\log\left(e\left(1+\frac{n}{S-1}\right)\right)$. Applying \Cref{lem:trans} and \Cref{lem:var-pq} to the terms appearing in \Cref{eq:empirical} gives:
    \begin{align}
        \EE_{s'\sim p_\mu(\cdot\mid s,a)}[d(s',\overline{s}_{s,a})]&\le \EE_{s'\sim \hat{p}_\mu(\cdot\mid s,a)}[d(s',\overline{s}_{s,a})]+\sqrt{2\mathrm{Var}_{\hat{p}_\mu}\left[d(s',\overline{s}_{s,a})\right]\frac{g(S,A,n,\alpha)}{n}}\nonumber\\
        &\quad +\frac{2d_s g(S,A,n,\alpha)}{3n} \tag{\Cref{lem:trans}} \\
        &\le \EE_{s'\sim \hat{p}_\mu(\cdot\mid s,a)}[d(s',\overline{s}_{s,a})]+\sqrt{\frac{2d_s^2 g(S,A,n,\alpha)}{n}}+\frac{2d_s g(S,A,n,\alpha)}{3n}\tag{$\mathrm{Var}_p[f]\le \EE_{p}[f^2]$},
    \end{align}
    and
    \begin{align}
        \sqrt{2\mathrm{Var}_{p_\mu}(d(s',\overline{s}_{s,a}))\varepsilon}&\le 2\sqrt{\mathrm{Var}_{\hat{p}_\mu}(d(s',\overline{s}_{s,a}))\varepsilon+2 d_s^2\varepsilon \frac{g(S,A,n,\alpha)}{n}}\tag{\Cref{lem:var-pq}}\\
        &\le 2\sqrt{\mathrm{Var}_{\hat{p}_\mu}(d(s',\overline{s}_{s,a}))\varepsilon}+2\sqrt{2 d_s^2\varepsilon \frac{g(S,A,n,\alpha)}{n}}\tag{$\sqrt{x+y}\le\sqrt{x}+\sqrt{y}$}.
    \end{align}
    Substituting these into \Cref{eq:empirical} yields that with probability at least $1-\alpha$, for any $(s,a)$ and any $\xi\in\Xi$, the following inequality holds:
    \begin{align}
        D_W(p_\xi,p^\star)(s,a)&\le 2\EE_{s'\sim \hat{p}_\mu(\cdot\mid s,a)}[d(s',\overline{s}_{s,a})]+2\sqrt{\frac{2d_s^2 g(S,A,n,\alpha)}{n}}+\frac{4d_s g(S,A,n,\alpha)}{3n}\nonumber\\
        &\quad +4\sqrt{\mathrm{Var}_{\hat{p}_\mu}(d(s',\overline{s}_{s,a}))\varepsilon}+4\sqrt{2 d_s^2\varepsilon \frac{g(S,A,n,\alpha)}{n}}+\frac{4}{3}d_s\cdot \varepsilon\nonumber\\
        &\le \frac{2}{c_3}(\EE_{s'\sim \hat{p}_\mu(\cdot\mid s,a)}[d(s',\overline{s}_{s,a})])^2+2\sqrt{\frac{2d_s^2 g(S,A,n,\alpha)}{n}}+\frac{4d_s g(S,A,n,\alpha)}{3n}\nonumber\\
        &\quad +4\sqrt{\mathrm{Var}_{\hat{p}_\mu}(d(s',\overline{s}_{s,a}))\varepsilon}+4\sqrt{2 d_s^2\varepsilon \frac{g(S,A,n,\alpha)}{n}}+\frac{4}{3}d_s\cdot \varepsilon\tag{\Cref{assumption:regularization}}\\
        &\le \frac{2}{c_3}(\EE_{s'\sim \hat{p}_\mu(\cdot\mid s,a)}[d(s',\overline{s}_{s,a})])^2+2\sqrt{\frac{2d_s^2 g(S,A,n,\alpha)}{n}}+\frac{4d_s g(S,A,n,\alpha)}{3n}\nonumber\\
        &\quad +\frac{2}{c_3}\mathrm{Var}_{\hat{p}_\mu}(d(s',\overline{s}_{s,a}))+4\sqrt{2 d_s^2\varepsilon \frac{g(S,A,n,\alpha)}{n}}+(\frac{4}{3}d_s+2c_3)\cdot \varepsilon\tag{Young's inequality}\\
        &=\frac{2}{c_3}\upsilon(s,a)+2\sqrt{\frac{2d_s^2 g(S,A,n,\alpha)}{n}}+\frac{4d_s g(S,A,n,\alpha)}{3n}\nonumber\\
        &\quad +4\sqrt{2 d_s^2\varepsilon \frac{g(S,A,n,\alpha)}{n}}+(\frac{4}{3}d_s+2c_3)\cdot \varepsilon,
    \end{align}
    where the last equality utilizes that $\EE[f^2]=(\EE[f])^2+\mathrm{Var}[f]$ and the fact that
    \begin{equation*}
        \upsilon(s, a) = \sum_{j=1}^{\operatorname{dim}(\cS)} \mathrm{Var}\left( s_{1,j},\ldots,s_{n,j} \right)=\EE_{s'\sim \hat{p}_\mu(\cdot\mid s,a)}\bigg[[d(s',\overline{s}_{s,a})]^2\bigg]
    \end{equation*}
    This ends the proof.
\end{proof}

\newpage

\section{Additional Sim-to-Sim Ablations}
\label{sec:additional-ablations}
We conduct further empirical analysis of \algname{SPiDR}, focusing on the following key aspects: 
\begin{enumerate*}[label=\textbf{(\roman*)}]
    \item sensitivity to the penalty parameter $\lambda$,
    \item further analysis of standard domain randomization,
    \item empirical validation of \Cref{lem:simulation} and,
    \item behavior of the uncertainty approximation $\upsilon(s, a)$
\end{enumerate*}
Through these ablations, we highlight \algname{SPiDR}'s robustness to the choice of $\lambda$, its efficiency in scaling to larger domain ensembles, and its principled handling of uncertainty in challenging state-action regions.

\begin{figure}[h]
    \centering
    \includegraphics{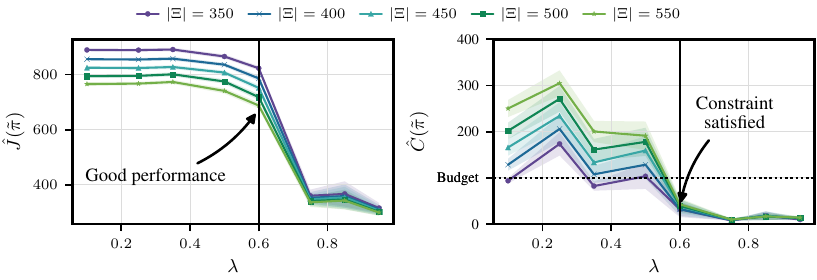}
    \caption{\algname{SPiDR}'s performance under different values of motor gear parameter. One choice of $\lambda$ consistently maintains constraint satisfaction.}
    \label{fig:tune-lambda}
\end{figure}

\paragraph{How robust is \algname{SPiDR} to the choice of $\boldsymbol{\lambda}$?}
We study the performance of \algname{SPiDR} across varying values of $\lambda$ and under different magnitudes of distribution shifts in the CartpoleBalance task. To this end, we vary the magnitude of the actuator's gear parameter, denoted as $\lvert\Xi\rvert$, with a slight abuse of notation. We ablate $\lambda \in \{0.85, 0.95, 0.75, 0.6 , 0.5 , 0.25, 0.35, 0.1 , 0\}$ and $\lvert\Xi\rvert$ across $\{350, 400, 500, 500\}$. We present our results in \Cref{fig:tune-lambda}, where we compare the objective and constraint for different $\lambda$ values. Notably, for $\lambda = 0.6$, 
while some performance is sacrificed in the objective, depending on the magnitude of $\lvert\Xi\rvert$, the constraint is satisfied across all settings of $\lvert\Xi\rvert$.

\begin{figure}[h]
    \centering
    \includegraphics{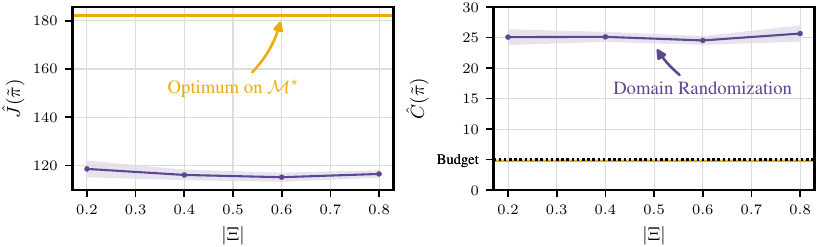}
    \caption{Constraint and objective of domain randomization when varying the size of the car's throttle parameter. Constraints are violated even when increasing the range of paramteres.}
    \label{fig:dr-ablation}
\end{figure}

\paragraph{Can domain randomization still be safe if we enlarge the set $\boldsymbol{\Xi}$?}
We study the dependence of standard domain randomization on the size of the set $\Xi$, presenting an example where even significantly increasing this set, and therefore the ``diversity'' of training environments, fails to enable safe transfer. To this end, we use the simulated RaceCar environment and vary the lower and upper limits of the car's throttle parameter. This parameter controls how fast can the car drive, directly relating to its ability to operate safely upon deployment. Specifically, we use the following values $\{(0.4, 0.6), (0.3, 0.7), (0.2, 0.8), (0.1, 0.9)\}$, where each tuple defines the minimum and maximum throttle values. The default range used in our experiments (including sim-to-real experiments) is $(0.4, 0.6)$. As in the previous experiment, we slightly abuse notation and refer to the diameter of $\Xi$ as $\lvert\Xi\rvert$, computed as the difference between the upper and lower bounds. We report our results in \Cref{fig:dr-ablation},
demonstrating that for this task, safety upon transfer is not maintained, even when using a wide range of parameters with domain randomization. More generally, we argue that for some problems, simply enlarging $\Xi$ could generally improve safety, however when safety must be ensured, pessimism is necessary.

\begin{figure}[h]
    \centering
    \includegraphics{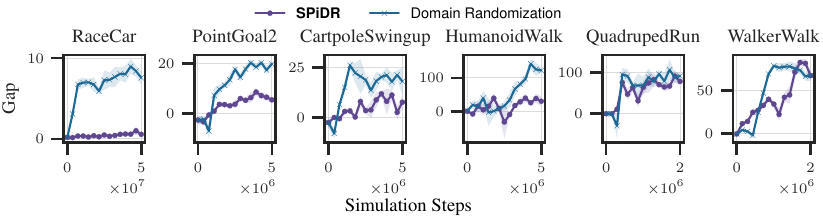}
    \caption{Constraint performance gap between training and test (lower is better). Domain randomization underestimates the constraint on the test tasks.}
    \label{fig:sim-to-sim-gap}
\end{figure}

\paragraph{Can we quantify constraint underestimation?}
We investigate the degree of constraint underestimation from \Cref{lem:simulation}. To this end, we measure the difference in the constraint measured in the training environment against the evaluation environment. Concretely, we collect empirical estimates of $C_{\star}(\tilde{\pi})$ and $\EE_{\xi\sim \mu}C_{\hat{p}_\xi}(\tilde{\pi})$ and report $C_{\star}(\pi) - \EE_{\xi\sim \mu}C_{\hat{p}_\xi}(\tilde{\pi})$ for \algname{SPiDR} and standard domain randomization in \Cref{fig:sim-to-sim-gap}. As shown, in all tasks, compared to \algname{SPiDR}, standard domain randomization underestimates the cost, leading to unsafe behavior on the test tasks (cf. \Cref{fig:simulated-costs}).

\begin{figure}[h]
    \centering
    \includegraphics{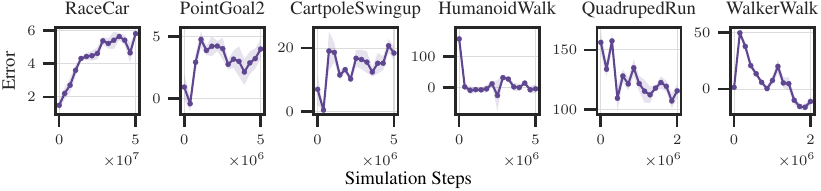}
    \caption{Our estimator $\upsilon(s, a)$ upper bounds the model discrepancy.}
    \label{fig:sim-to-sim-estimator}
\end{figure}

\paragraph{Does $\boldsymbol{\upsilon(s, a)}$ upper-bounds the discrepancy in practice?}
We continue our previous study and demonstrate that $\EE_{(s,a)\sim d_{\hat{p}_\xi,\pi}} [\lambda\upsilon(s, a)] - \left|C_{\star}(\pi) - \EE_{\xi\sim \mu}C_{\hat{p}_\xi}(\pi)\right| \gtrsim 0$, namely, that our estimate $\upsilon(\cdot, \cdot)$ can be used to sufficiently penalize the cost. In \Cref{fig:sim-to-sim-estimator} we report this error across all six simulated tasks.
As shown, in all tasks, our approximation error is positive, suggesting that $\lambda\upsilon$ indeed upper-bounds the gap in constraint. Notably, in the HumanoidWalk environment, the error of $\lambda \upsilon(s, a)$ w.r.t. the constraint gap is close to zero. This result is in line with the performance-safety tradeoff shown in \Cref{fig:simulated}, were \algname{SPiDR} is both safe and achieving strong performance compared to the best solution on $\realmdp$.

\begin{figure}[h]
    \centering
    \includegraphics{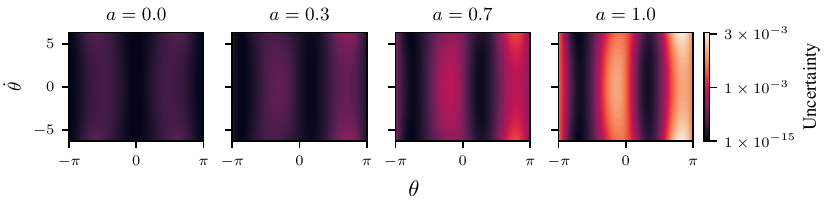}
    \caption{Uncertainty $\upsilon(s, a)$ across angular states for actions $a \in \{0.0, 0.3, 0.7, 1.0\}$. Uncertainty increases with action magnitude and is highest near the upright position ($\theta \approx \pi$), highlighting states where the simulator's predictions are less reliable.}
    \label{fig:cartpole-disagreement}
\end{figure}

\paragraph{How does $\boldsymbol{\upsilon(s, a)}$ vary across states?}
We analyze $\upsilon(s, a)$ measured across different states and actions in the CartpoleSwingup environment. The state space consists of the linear cart position and velocity, along with the angular position ($\theta$) and angular velocity ($\dot{\theta}$) of the pole. The continuous action space lies in $[-1, 1]$, and due to the symmetry of the system, we restrict our analysis to $a \in [0, 1]$. \Cref{fig:cartpole-disagreement} visualizes the uncertainty over the angular position and velocity dimensions for representative actions $a \in \{0.0, 0.3, 0.7, 1.0\}$. We note the following observations:
\begin{enumerate*}[label=\textbf{(\roman*)}]
    \item uncertainty generally increases with action magnitude, and, 
    \item for $a \neq 0$, uncertainty peaks when the pole is near the upright position ($\theta \approx \pi$), which corresponds to an inherently unstable equilibrium.
\end{enumerate*}
This observation aligns with intuition: near the unstable upright position, small perturbations can lead to significantly different outcomes, making these regions harder to model. From a safety perspective, this analysis motivates penalizing high-uncertainty regions during policy learning, particularly when deploying on real systems where model inaccuracies may have significant consequences.

\newpage

\section{Picking $\boldsymbol{\lambda}$ in Practice}
\label{sec:choosing-lambda}
We demonstrate our ablations on $\lambda$ when deploying \algname{SPiDR} in real, providing guidance on how to pick it effectively in practice. We describe below the procedure we use in our sim-to-real experiments for selecting $\lambda$.

\begin{enumerate}[label=Step \arabic*:, leftmargin=3.3em]
  \renewcommand{\theenumi}{Step~\arabic{enumi}} 
  \item \textbf{Estimate the magnitude of $\boldsymbol{\upsilon(\cdot, \cdot)}$.}
Evaluate and record $\upsilon(\cdot, \cdot)$ in training, aggregated over state-action pairs and across time. This estimate is not required to be precise, its purpose is to determine the general order of magnitude. This can be done purely in simulation.
  \item \textbf{Select an appropriate range for $\boldsymbol{\lambda}$.}  
Choose a candidate range of $\lambda$ values such that $\nicefrac{1}{\upsilon} \approx 1$. This heuristic ensures that the penalty term has roughly the same magnitude of cost function, assuming $c_{\text{max}} \approx 1$.
    \item \textbf{Iteratively refine $\boldsymbol{\lambda}$.}  \label{step:refine-lambda}
On the real system, begin testing with the largest value in the selected range and iteratively decrease it until the constraint satisfaction on the real system closely matches the desired budget.
\end{enumerate}

\begin{figure}[h]
    \centering
    \begin{subfigure}[b]{0.49\textwidth}
        \centering
        \includegraphics[width=\textwidth]{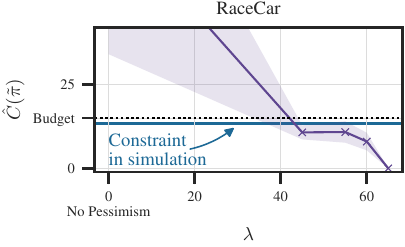}
        \label{fig:rccar-lambda-ablation}
    \end{subfigure}
    \hfill
    \begin{subfigure}[b]{0.49\textwidth}i
        \centering
        \includegraphics[width=\textwidth]{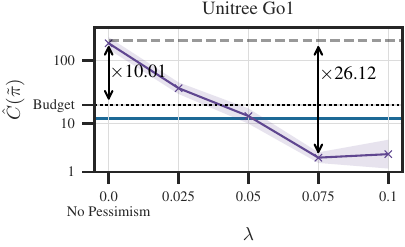}
        \label{fig:go1-lambda-ablation}
    \end{subfigure}
    \vspace{-0.5cm}
    \caption{Safety performance under different $\lambda$ values in real-world robotic tasks. The blue horizontal line represents the value of the constraint measured \emph{in simulation} under $\lambda = 0$. \algname{SPiDR} satisfies the constraints on both of the physical systems.}
    \label{fig:sim-to-real}
\end{figure}

In \Cref{fig:sim-to-real} we report the performance of \algname{SPiDR} across different values of $\lambda$.
Importantly, for a large enough initial choice of $\lambda$, which can be obtained in practice using prior domain knowledge (e.g. degree of of simulator fidelity), the very first deployment of \algname{SPiDR} is safe. In \ref{step:refine-lambda}, online data is used to improve \emph{performance}, while maintaining safety. With this procedure, safety is maintained zero-shot, while performance can be improved with online rollouts, in line with our safety guarantee in \Cref{sec:theoretical-things}.

\section{Learning Curves for Simulated Environments}
\label{sec:learning-curves}
\begin{figure}[h]
    \centering
    \includegraphics{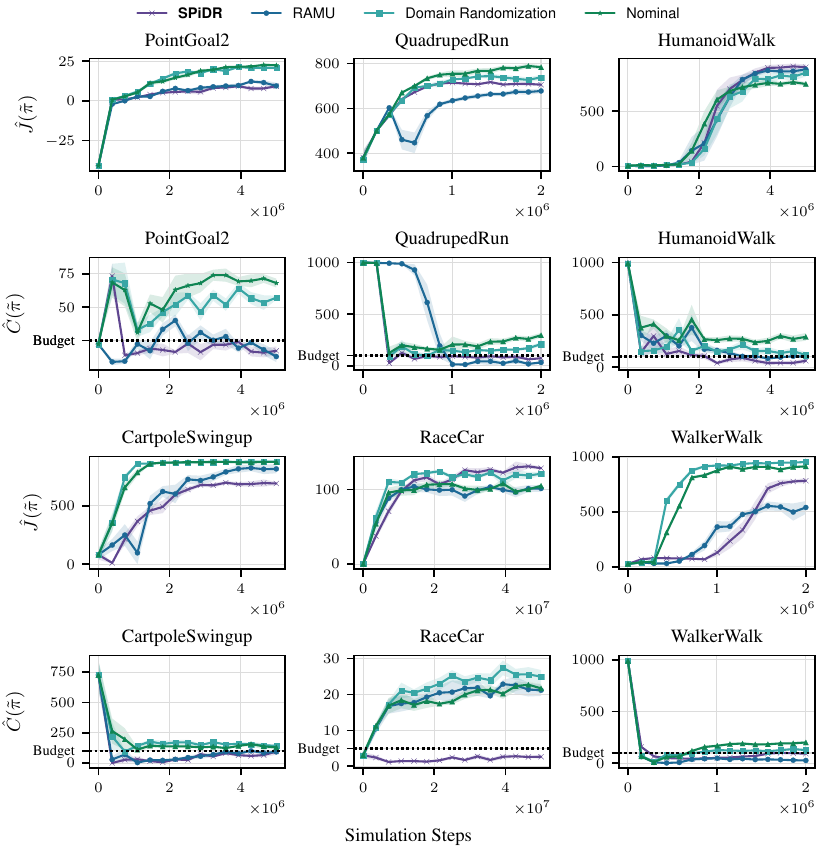}
    \caption{Learning curves used in \Cref{sec:sim-to-sim}. \algname{SPiDR} consistently satisfies the constraints while maintaining good performance on the objective. \algname{Domain Randomization} and \algname{Nominal} fail to satisfy the constraints.}
    \label{fig:learning-curves}
\end{figure}

In \Cref{fig:learning-curves} we provide the full learning curves of the experiment trials used for \Cref{sec:sim-to-sim}, including the standard error intervals across five random seeds.

\paragraph{Simulating the sim-to-sim gap.}
We simulate the sim-to-sim gap as follows. In PointGoal2 and the RWRL tasks, we follow a similar approach to \citet{risk-averse} and introduce in evaluation an additional dynamics parameter (e.g., mass or motor gains) that is not encountered during training. In the RaceCar environment, the agent is trained using a simplified bicycle model, but evaluated on a more realistic variant that incorporates tire forces and detailed motor dynamics. Further details on the tasks and their sim-to-sim gap design can be found in \Cref{sec:safety-gym,sec:rwrl,sec:racecar}. 

\newpage

\section{Locomotion Experiments with Unitree Go1}
\label{sec:go1-locomotion}

We train policies using the FlatTerrainGo1Joystick environment from MuJoCo Playground \citep{zakka2025mujocoplayground}. In this environment, the agent is tasked with following randomly sampled velocity commands in the forward, lateral, and yaw directions. Excessive joint motion can cause the legs to self-collide, even when joint limits are not reached. These collisions can lead to falls or serious hardware damage.
\paragraph{Constraints.}
To prevent these joint limit violations we define a cost function to measure the number of joint violations. The cost function is defined by the indicator function of any joint being outside the soft limit. More formally, the cost is defined as
\begin{equation*}
    c(s, a) \triangleq 
    \begin{cases}
    1 & \text{if any of } j \in \mathcal{J} \text{ such that } q_j > 0.75 \cdot q^{\max}_j \text{ or } q_j < 0.75 \cdot q^{\min}_j  \\
    0 & \text{otherwise}
    \end{cases}    
\end{equation*}
where $\mathcal{J}$ is the set of joint indices, $q_j$ is the joint angle and $q^{\max}_j, q^{\min}_j$ are the angle limits of joint angle $j$.
The soft factor limit of 0.75 is applied meaning that if a joint enters the outermost 25\% of its feasible range of motion, the cost is set to 1. We choose this constraint since joint-position measurements are accurate and reliable on the real system, without relying on indirect filtering/estimation methods.
\paragraph{Training in simulation.}
Each policy is trained for roughly one hour on an NVIDIA RTX4090 GPU. For both SAC and PPO, we train policies with a primal-dual solver, using different values of $\lambda$ to penalize the uncertainty. Each $\lambda$ value is trained across five different random seeds.

\paragraph{Command distribution.}
Target commands are uniformly sampled from the ranges $[ \pm 0.45, \pm 0.2, \pm 1.3 ]$, corresponding to the forward velocity, lateral velocity, and yaw rate, respectively. Each sampled command is applied for a fixed number of 1000 control steps. After this period, the command is reversed, perturbed with additive noise, and then reapplied for the same duration, after which a new command is sampled from the same distribution as stated above.

\paragraph{Real-world evaluation.}
We evaluate the trained policies on the Unitree Go1 quadruped robot. To ensure that all policies are evaluated under identical conditions, the same sequences of commands are sampled for each episode. Every policy is tested across 10 independent trials. We provide \texttt{.onnx} files for the policies used in this experiment, as well as the code to port simulation-trained policies to \texttt{.onnx} format in \href{https://github.com/yardenas/safe-learning}{the following link}.

\begin{figure}[h]
    \centering
    \begin{tabular}{c @{\hskip 5pt} c @{\hskip 5pt} c @{\hskip 5pt} c @{\hskip 5pt} c}
    \rotatebox{90}{\algname{RAMU}} &
        \begin{tikzpicture}
            \node[anchor=south west, inner sep=0] (image1) at (0, 0) 
                {\includegraphics[width=0.22\textwidth, clip]{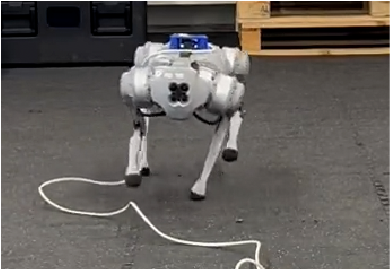}};
        \end{tikzpicture} &
        \begin{tikzpicture}
            \node[anchor=south west, inner sep=0] (image2) at (0, 0) 
                {\includegraphics[width=0.22\textwidth, clip]{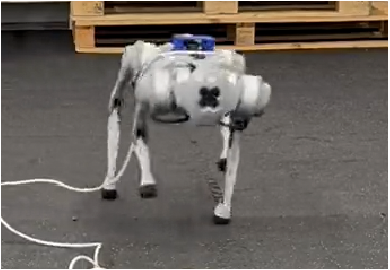}};
        \end{tikzpicture} &
        \begin{tikzpicture}
            \node[anchor=south west, inner sep=0] (image3) at (0, 0) 
                {\includegraphics[width=0.22\textwidth, clip]{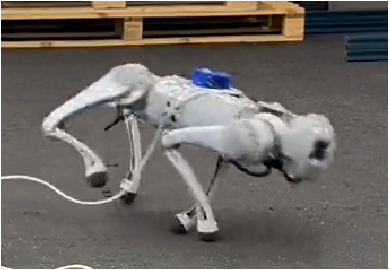}};
        \end{tikzpicture} &
        \begin{tikzpicture}
            \node[anchor=south west, inner sep=0] (image4) at (0, 0) 
                {\includegraphics[width=0.22\textwidth, clip]{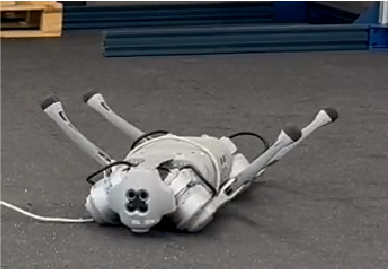}};
        \end{tikzpicture} \\
        \rotatebox{90}{\algname{SPiDR}} &
        \begin{tikzpicture}
            \node[anchor=south west, inner sep=0] (image5) at (0, 0) 
                {\includegraphics[width=0.22\textwidth, clip]{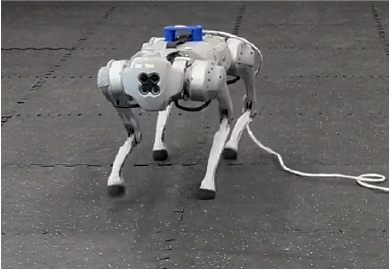}};
        \end{tikzpicture} &
        \begin{tikzpicture}
            \node[anchor=south west, inner sep=0] (image6) at (0, 0) 
                {\includegraphics[width=0.22\textwidth, clip]{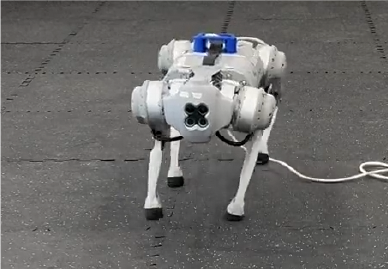}};
        \end{tikzpicture} &
        \begin{tikzpicture}
            \node[anchor=south west, inner sep=0] (image7) at (0, 0) 
                {\includegraphics[width=0.22\textwidth, clip]{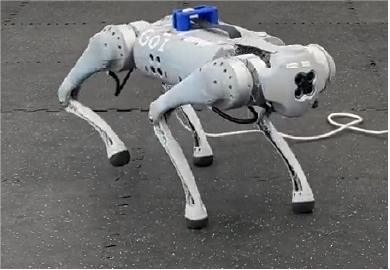}};
        \end{tikzpicture} &
        \begin{tikzpicture}
            \node[anchor=south west, inner sep=0] (image8) at (0, 0) 
                {\includegraphics[width=0.22\textwidth, clip]{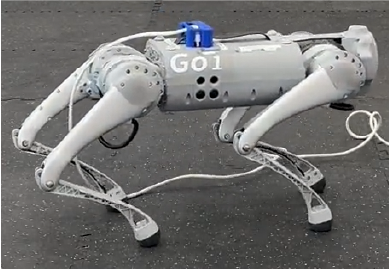}};
        \end{tikzpicture} \\
    \end{tabular}
    \caption{Comparison of \algname{RAMU} and \algname{SPiDR} on the Unitree Go1 robot. \algname{SPiDR} follows the given commadns without falling.}
    \label{fig:go1-ramu-comparison}
\end{figure}

\paragraph{Comparison with \algname{RAMU}.}
In \Cref{fig:go1-ramu-comparison} we demonstrate a comparison of \algname{SPiDR} and \algname{RAMU}, conditioned on the same sequence of commands. We note that \algname{RAMU} uses only a single ``nominal'' environment, together with robust estimation value function estimation as means for robust transfer. We believe that \algname{RAMU} fails to follow the commands mainly because using only the nominal environment, without incorporating domain randomization, might not be sufficient for robust sim-to-real transfer. We note that conceptually \algname{RAMU} and domain randomization are not mutually exclusive, therefore \algname{RAMU} could in principle transfer better if used with domain randomization. However, combining the two goes beyond the proposed algorithmic solution and official implementation of \algname{RAMU}, and therefore not considered in this experiment. Despite that, we believe that combining the two approach is an interesting direction for future work.

\section{RaceCar Experiments}
\label{sec:racecar-real}
In this task we implement the simulated RaceCar environment on a real remote-controlled car, illustrated in \Cref{fig:hardware-demo}. Please see \Cref{sec:racecar} for further details about the reward and cost functions.

\paragraph{Real-world implementation.} We measure the position and orientation of the car using a motion-capture system. Velocities are estimated using a first-order low pass filter. These measurements are sufficient to recover the full state of the system as the goal and obstacle positions are fixed. All trajectory measurements start from the roughly same initial position in the world frame. The goal position is at the origin. Further details, regarding the parameters used for domain randomization and training hyper-parameters, can be found in \href{https://github.com/yardenas/safe-learning}{our open-source implementation}.

\begin{figure}[h!]
    \centering
    \includegraphics{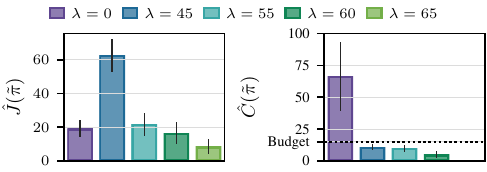}
    \caption{Safety and objective performance for $\lambda \in \{0, 45, 55, 60, 65\}$ on the real system. We report the mean and standard error across five seeds. \algname{SPiDR} transfers safely to the real system while solving the task, i.e., reaching to the goal position.}
    \label{fig:rccar-sim-to-real}
\end{figure}

\paragraph{Additional results.}
In \Cref{fig:rccar-sim-to-real} we provide the objective and constraint measured on the real system in this experiment. As shown, for $\lambda = 45$, \algname{SPiDR} is able to satisfy the constraint, while finding a performant policy.

\begin{figure}[h!]
    \centering
    \includegraphics{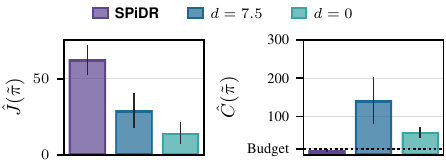}
    \caption{Comparison of \algname{SPiDR} with constraint tightening on the real RaceCar. Constraint tightening fails to satisfy the constraint.}
    \label{fig:constraint-tightening}
\end{figure}

\paragraph{Comparison with constraint tightening.}
We additionally compare \algname{SPiDR} with a simple baseline that tightens the constraint budget $d$ in simulation. To this end, we evaluate ``constraint tightening'' on the real-world RaceCar, ablating it when using budgets $d \in \{0, 7.5\}$ when training in simulation. We present our results in \Cref{fig:constraint-tightening}. As shown, while reducing the budget $d$ in simulation indeed reduces the degree of accumulated costs on the real system, it might still not be sufficient to maintain safety, opposed to \algname{SPiDR}.
While this approach is even simpler than \algname{SPiDR}, there are two main challenges with it. First, severe tightening may yield a ``zero budget'' training regime that CMDP solvers like CRPO and primal-dual methods struggle to solve in practice \citep{huang2023safedreamer,he2023autocost}. This behavior is observerd in \Cref{fig:constraint-tightening}; even though the budget is set to $d = 0$, the primal-dual CMDP solver we use fails to converge to a constraint satisfying solution.
Second, as hinted, reducing the budget is equivalent to finding a uniform upper-bound to the penalty term in \Cref{eq:penalty-term}, that does not depend on state-actions. Therefore, in order to achieve the same level of penalty required to satisfy the constraint, this for of ``uniform pessimism'' can degrade performance by being over-pessimistic in those states where the uncertainty about the sim-to-real gap might in fact be low, effectively penalizing the ``wrong'' states. We visualize such state-action-dependent uncertainty in \Cref{fig:cartpole-disagreement}.

\section{Additional Experiments with PPO}
\label{sec:ppo}
\begin{figure}[h!]
    \centering
    \includegraphics{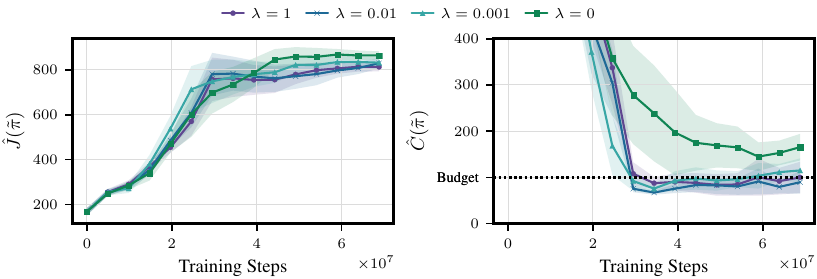}
    \caption{PPO with SauteRL as a CMDP solver. Domain randomization fails to satisfy the constraints, while for $\lambda \ge 0.01$ constraints are satisfied with minor performance drop.}
    \label{fig:ppo-curves}
\end{figure}

To further study \algname{SPiDR}'s performance with CMDP solvers that utilize policy gradients, we use PPO \citep{schulman2017proximal}. PPO is a common choice used in many robotics tasks \citep[for instance, see][]{joonho2020}. We use SauteRL  as a CMDP solver \citep{sootla2022saute}. SuateRL augments the state with a counter of the online accumulated cost, and penalizes the reward once this counter exceeds the budget. We ablate $\lambda \in \{0, 0.001, 0.01, 1\}$ and report the performance on the CartpoleBalance task from RWRL in \Cref{fig:ppo-curves}.
As shown, the constraints are violated for $\lambda = 0$, corresponding to standard domain randomization. In contrast, the constraints are satisfied as $\lambda$ increases. We note that, while SauteRL is designed for constraint satisfaction with high probability, instead of bounding the expectation of the cumulative costs, and thus in principle it adds additional conservatism, this might still not guarantee safe transfer under modeling mismatches. \Cref{fig:ppo-curves} demonstrates that \algname{SPiDR} mitigates this issue.

\section{Safety Gym}
\label{sec:safety-gym}
To compare policies in environments from OpenAI Safety Gym by \citet{Ray2019}, we port the PointGoal2 environment from standard MuJoCo to MJX (MuJoCo XLA). MJX is a JAX-compiled MuJoCo backend that is tightly integrated with Brax \citep{freeman2021brax}. This port enables us to massively parallelize training by collecting trajectories from thousands of environments in parallel on a single GPU, accelerating training by several orders of magnitude compared to the original implementation. This contribution is of independent interest to the safe RL community. Extending our work to additional Safety Gym environments is left for future work.

\begin{figure}[h]
\centering
\begin{minipage}{0.45\linewidth}
    \centering
    \includegraphics[width=\linewidth]{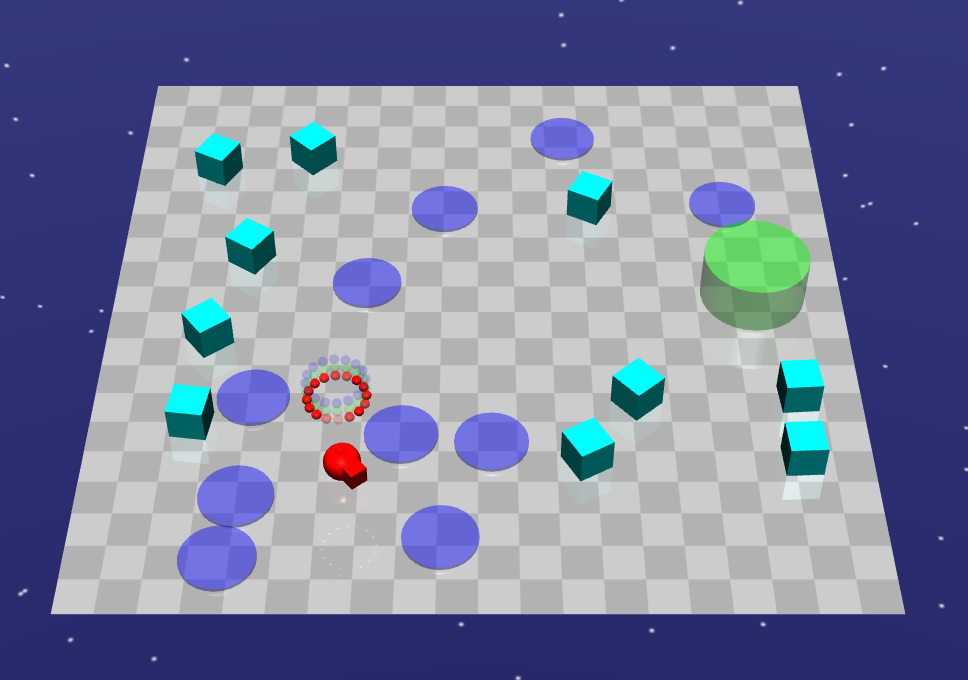}
    \caption{Visualization of a random initialization of the PointGoal2 environment. The red pointmass is the agent, the green transparent cylinder is the goal, the cyan boxes are vases and the blue circles are hazard zones.}
    \label{fig:safety-gym}
\end{minipage}\hfill
\begin{minipage}{0.52\linewidth}
    \centering
    \vspace{-0.4cm}
    {\renewcommand{\arraystretch}{1.5}
    \captionof{table}{Domain randomization parameters and ranges used during training and evaluation. $+$ and $\times$ denote additive and multiplicative terms respectively. By fixing the damping and mass parameters in training, but not in evaluation, we simulate both lack of knowledge of $\xi^\star$, but also modeling mismatch due to imperfect simulation.}
    \label{tab:domain-randomization}
    \begin{tabular}{l l l}
    \hline\hline 
    \textbf{Parameter}   & \textbf{Train} & \textbf{Eval} \\
    \hline  
    Damping (x, y) $\times$       & 1.0 (fixed)          & [0.6, 1.0]            \\
    Damping (z) $\times$         & 1.0 (fixed)          & [0.7, 1.0]            \\
    Gear (x) $+$             & [–0.2, 0.2]           & [0, 0.1]              \\
    Gear (z) $+$            & [–0.1, 0.1]           & [0, 0.05]             \\
    Mass $\times$                & 1.0 (fixed)          & [1.0, 1.05]           \\
    \hline  
    \end{tabular}
    }
\end{minipage}
\end{figure}

\paragraph{PointGoal2 environment.} In this environment, the agent must navigate to a target location while avoiding hazards which include free-moving vases and designated hazard zones. The environment is depicted in \Cref{fig:safety-gym}. The initial positions of the agent, goal, vases, and hazards are randomized at the beginning of training. The environment reward is defined as the change in Euclidean distance to the goal between successive steps
\begin{equation*}
    r_t(s_t, a_t) \triangleq d_{t-1} - d_t + \1[d_t \leq \epsilon],
\end{equation*}
where $d_t = \|\mathbf{x}_t - \mathbf{x}_{\text{goal}}\|_2$ is the Euclidean distance from the robot to the goal. The term $\1[d_t \leq \epsilon]$ is an indicator function that gives a reward bonus when the agent reaches the goal, i.e., when it is within $\epsilon = 0.3$ of the center of the goal. The goal position is resampled to another free position in the environment once reached.

A cost of 1 is incurred when the agent collides with a vase $v$, when one of the vases crosses a linear velocity threshold (after collision), or when the agent is inside a hazard zone $h$: 
\begin{equation*}
    c_t(s_t, a_t) \triangleq \1[\exists v\in V:\text{collides}(\mathbf{x}_t, \mathbf{x}_{v})] + \1[\exists v\in V:\mathbf{x}'_v \geq \gamma] + \1[\exists h \in H: d_t \leq \rho],
\end{equation*}
where  $\gamma=5e^{-2}$ and $\rho=0.2$. Please see the implementation of \citet{Ray2019} for more specific details and \href{https://github.com/yardenas/safe-learning}{our open-source implementation}.

\paragraph{Sim-to-sim gap.}
During evaluation, we uniformly sample three key parameters of the environment's dynamics: multiplicative joint damping and mass factors and additive values for the actuator gear ratio. The precise training and evaluation ranges are given in \cref{tab:domain-randomization}. Note that the $z$-joint is a hinge joint that allows the agent to rotate around the $z$-axis and the $x$-joint is a slide joint that allows translation in the $xy$-plane.

\section{RWRL Benchmark}
\label{sec:rwrl}
We evaluate \algname{SPiDR} on four robotic tasks using the RWRL benchmark \citep{dulacarnold2020realworldrlempirical}, which adds safety constraints and distribution shifts to DeepMind Control suite tasks. We build on MuJoCo Playground \citep{zakka2025mujocoplayground}, an MJX-based reimplementation that enables faster, parallelized training, by incorporating RWRL's modifications. See our open-source implementation for details.
\begin{figure}[h]
    \centering
    \begin{tabular}{@{\hskip 5pt} c @{\hskip 5pt} c @{\hskip 5pt} c @{\hskip 5pt} c @{\hskip 5pt} c}
        \includegraphics[height=70pt, trim={2.5pt 2.5pt 2.5pt 2.5pt 2.5pt}, clip]{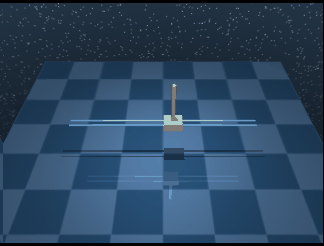} &
        \includegraphics[height=70pt, trim={2.5pt 2.5pt 2.5pt 2.5pt 2.5pt}, clip]{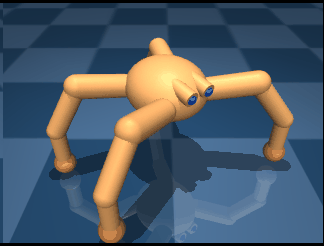} &
        \includegraphics[height=70pt, trim={2.5pt 2.5pt 2.5pt 2.5pt 2.5pt}, clip]{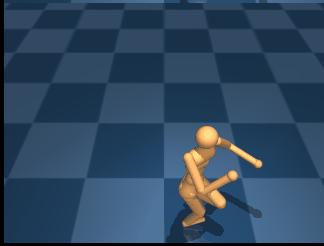} &
        \includegraphics[height=70pt, trim={2.5pt 2.5pt 2.5pt 2.5pt 2.5pt}, clip]{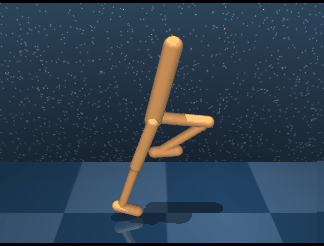}
    \end{tabular}
    \caption{RWRL tasks.}
    \vspace{-0.5cm}
    \label{fig:rwrl-demo}
\end{figure}

\paragraph{Constraints.}
We use the joint position limits constraint for HumanoidWalk and QuadrupedRun, joint velocity limits for WalkerWalk, and slider position limits for CartpoleSwingup. These are the standard constraints proposed by \citet{dulacarnold2020realworldrlempirical}.

\paragraph{Sim-to-sim gap.}
We follow a similar experimental setup as \citet{risk-averse} and introduce additional variability in the evaluation task to simulate modeling mismatches. In \Cref{tab:domain-randomization-multitask} we provide the specific parameters we perturb for the RWRL environments.
\begin{table}[h!]
\caption{Domain randomization parameters and ranges used during training and evaluation across tasks from RWLR. $+$ and $\times$ denote additive and multiplicative terms respectively.}
\vspace{0.1cm}
\centering
{\renewcommand{\arraystretch}{1.5} 
\begin{tabular}{l l l}
\hline\hline 
\textbf{Parameter}   & \textbf{Train} & \textbf{Eval} \\
\hline  
\multicolumn{3}{c}{\textbf{CartpoleSwingup}} \\
\hline
Pole Length $+$          & [0.0, 0.0]             & [–0.25, 0.25]         \\
Gear $+$                & [0.0, 5.0]             & [0.0, 5.0]            \\
\hline
\multicolumn{3}{c}{\textbf{QuadrupedRun}} \\
\hline
Torso Density $+$    & [0.0, 0.0]             & [–300.0, 300.0]       \\
Friction $+$            & [0.0, 0.0]             & [–0.95, 0.0]          \\
Lift Gear $\times$            & [0.75, 1.5]            & [0.75, 1.5]           \\
Yaw Gear $\times$            & [1.0, 1.0] (fixed)     & [0.5, 2.0]            \\
Extend Gear $\times$         & [0.75, 1.5]            & [0.75, 1.5]           \\
\hline
\multicolumn{3}{c}{\textbf{HumanoidWalk}} \\
\hline
Friction $+$            & [0.0, 0.0]             & [-0.05, 0.05]           \\
Hip Gear (x) $+$         & [-20., 20.]                 & [-20., 20.]               \\
Hip Gear (y) $+$         & [-20., 20.]                 & [-20., 20.]               \\
Hip Gear (z) $+$        & [-60., 60.]                 & [-60., 60.]               \\
Knee Gear $+$            & [0, 0]                 & [-40., 40.]               \\
\hline
\multicolumn{3}{c}{\textbf{WalkerWalk}} \\
\hline
Torso Length $+$        & [0.0, 0.0] (fixed)     & [–0.75, 0.75]         \\
Gear $+$                & [0.0, 20.0]            & [0.0, 20.0]           \\
\hline
\end{tabular}
}

\label{tab:domain-randomization-multitask}
\end{table}

\section{RaceCar Environment}
\label{sec:racecar}
\paragraph{Reward and cost.}
The reward at timestep $t$ is given by
\begin{equation*}
    r_t(s_t, a_t) \triangleq d_{t-1} - d_t + \1[d_t \leq \epsilon] - \lambda_c \|a_t\|_2 - \lambda_l \|a_t - a_{t-1}\|_2^2,
\end{equation*}
where $d_t = \|\mathbf{x}_t - \mathbf{x}_{\text{goal}}\|_2$ is the Euclidean distance to the goal, and $a_t \in \mathbb{R}^2$ denotes the action applied at time $t$ (consisting of steering and throttle). The term $\1[d_t \leq \epsilon]$ is an indicator function that gives a reward bonus when the agent is within $\epsilon = 0.3$ of the goal. The penalties $\lambda_c$ and $\lambda_l$ weight the control effort (magnitude of the action) and the change in action between consecutive timesteps, respectively.
The cost function at time $t$ is defined as
\begin{equation*}
    c_t \triangleq \sum_{i=1}^3 \1\left[\|x_t - p_i\| < \rho_i\right] E^k_t + \1[x_t \notin \mathcal{V}],
\end{equation*}

where $x_t \in \mathbb{R}^2$ is the agent's position, $p_i$ and $\rho_i$ are the position and radius of the $i$-th obstacle and $E^k_t$ is the kinetic energy of the car at time $t$, simulating a plastic collision between the car and obstacles. This choice of cost function allows us to penalize more severely collisions in which the car smashes into obstacles, as opposed to softly touching them. The second term penalizes the agent for leaving the valid area $\mathcal{V}$, which corresponds to a bounded rectangular arena.

\paragraph{Sim-to-sim gap.}
In the previous sim-to-sim environments, we model the sim-to-sim gap by introducing an auxiliary dynamics parameter (e.g., pendulum length) that is not observed during training. In contrast, in the RaceCar environment, the car dynamics in the training environments are governed by a semi-kinematic bicycle model that does not account for interactions between the tire and the ground. On the other hand, in evaluation, we use the dynamical bicycle model of \citet{kabzan2020amz}. We refer the reader to \citet{kabzan2020amz} for the detailed equations of motion as well as to \href{https://github.com/yardenas/safe-learning}{our open-source implementation} for more details.

\section{Additional Details on our CMDP Solvers}
\label{sec:cmdp-solvers}
\paragraph{CRPO.}
CRPO \citep{xu2021crpo} is a CMDP solver that uses a primal subgradient switching method to solve the constraint optimization problem. By doing so, CRPO is not tied to any specific RL algorithm and can be used with different policy search methods. For instance, \citet{xu2021crpo} use this technique with TRPO \citep{schulman2015trust} and \citet{risk-averse} use it with MPO \citet{abdolmaleki2018maximum}. In our experiments, we mainly use CRPO together with SAC \citep{haarnoja2019soft}. To apply CRPO with SAC, we learn a cost critic $Q_c(s, a)$, trained similarly to the reward critic but without the standard entropy term of SAC. To update the policy, we estimate the constraint with $\widehat{C}$, and if it exceeds the budget $d$, we switch from a reward-maximizing update to a cost-minimizing one. The procedure is detailed in \Cref{alg:crpo-sac}.

\begin{algorithm*}[h]
\caption{CRPO with SAC}
\algrenewcommand\algorithmicindent{1em}%
\label{alg:crpo-sac}
\begin{algorithmic}[1]
\State \textbf{Input:} Constraint budget $d$, replay buffer $\mathcal{D}$
\State \textbf{Initialize:} Policy $\pi_\theta$, reward critic $Q_r$, cost critic $Q_c$
\For{each environment interaction step}
    \State Collect action $a \sim \pi_\theta(\cdot \mid s)$
    \State Observe reward $r$, cost $c$, next state $s'$
    \State Store transition $(s, a, r, c, s')$ in $\mathcal{D}$
\EndFor
\For{each gradient update step}
    \State Sample batch $\{(s_i, a_i, r_i, c_i, s_i')\}_{i=1}^N$ from $\mathcal{D}$
    \State Update reward critic $Q_r$ using SAC critic loss
    \State Update cost critic $Q_c$ using TD error (no entropy bonus)
    \State Estimate constraint $\hat{C} \gets \frac{1}{N} \sum_{i=1}^N Q_c(s_i, \pi_\theta(s_i))$
    \Comment{Draw an action from $\pi_\theta$}
    \If{$\widehat{C} > d$}
        \Comment{Constraint violated: prioritize cost minimization}
        \State Update policy $\pi_\theta$ using gradient of $\widehat{C}$
    \Else
        \Comment{Constraint satisfied: optimize for reward}
        \State Update policy $\pi_\theta$ using SAC policy loss
    \EndIf
\EndFor
\State \Return Policy $\pi_\theta$
\end{algorithmic}
\end{algorithm*}

\paragraph{Primal-dual solvers.}
The primal-dual approach solves CMDPs by augmenting the standard loss used in PPO or SAC with a penalty term derived from the constraint using Lagrangian duality. Specifically, we define the Lagrangian as
\begin{equation*}
    \mathcal{L}(\theta, \lambda_{\text{PD}}) = J(\pi_\theta) - \lambda_{\text{PD}} \left(C(\pi_\theta) - d\right),
\end{equation*}
where $\lambda_{\text{PD}} \geq 0$ is the Lagrange multiplier. The goal is to find a saddle point of the Lagrangian by performing gradient descent on the policy parameters $\theta$ (primal variables) while performing gradient ascent on $\lambda_{\text{PD}}$ (dual variable) to enforce the constraint.

When integrating this method with SAC, we learn a cost critic together with the reward critic. When computing the policy loss, we evaluate the policy in a same way on the cost critic as we do in the reward critic to evaluate the constraint. The loss is computed via $\pi_\theta = \EE_{s, a}[\log \pi_\theta + Q_r(s, \pi() + \lambda_{\text{TD}} Q_c(s, \pi()]$. The same estimate $Q_c(s, \pi$ is used to update the dual
\begin{equation*}
    \lambda \leftarrow \left[\lambda + \eta_{\lambda_{\text{PD}}} \left(C(\pi_\theta) - d\right)\right]_+,
\end{equation*}
where $\eta_{\lambda_{\text{PD}}}$ is the dual learning rate.

For methods that rely on policy gradient estimators like PPO, we use a value-based estimate of the constraint via a learned cost value function $V_c(s)$, and compute advantage estimates accordingly. The policy update then follows the same principle, where the objective is penalized by the constraint estimate weighted by $\lambda_{\text{PD}}$, ensuring that constraint violations are actively discouraged during training.

More specific implementation details can be found in \href{https://github.com/yardenas/safe-learning}{our open-source implementation}.

\newpage

\section{Additional Related Work}
\label{sec:additional-related-works}
\paragraph{Safe reinforcement learning.}
\looseness=-1A prominent line of work addressing safety in RL utilizes constrained Markov decision processes. In CMDPs, agents optimize cumulative rewards while satisfying constraints on cumulative costs \citep{altman-constrainedMDP}. To solve CMDPs, two main algorithm categories are used: primal-dual and primal methods. Primal-dual approaches leverage the Lagrangian form, aiming to find a saddle-point of it. This is typically done by iteratively updating policy parameters and dual multipliers to balance rewards and constraints \citep{chow2018lyapunov,achiam2017constrained,yang2019projection,yang2022constrained}. On the other hand, primal methods bypass dual variables entirely, focusing on directly optimizing the primal problem by embedding constraints through gradient combination~\citep{xu2021crpo,gu2023pcrpo}. Other works use the CMDP formulation but propose alternative methods to satisfy the safety constraints. \citet{srinivasan2020learning} learn a safety actor-critic together with a behavior policy, \citet{thananjeyan2021recovery} use a recovery backup policy, \citet{bharadhwaj2020conservative} propose a conservative safety critic together with online rejection sampling of actions. Further discussions on alternative formulations of safety in RL are provided by \citet{garcia2015comprehensive,gu2022review,brunke2022safe}.
This work differs from standard CMDPs in that it focuses on addressing additional model mismatch within the CMDP framework.

\paragraph{Model uncertainty in reinforcement learning.}
\looseness=-1Various works address model uncertainty or model mismatch of the environment (e.g., reward function, dynamics, or task itself) during train- and test-time. This challenge is particularly relevant in sim-to-real transfer, which typically serves as a natural motivation for these works. Within this context, a line of research applies robust optimization to optimize worst-case performance, known as robust RL \citep{iyengar2005robust,xu2012distributionally,wolff2012robust,kaufman2013robust,tamar2014scaling,pinto2017robust,pattanaik2017robust,ho2018fast,tessler2019action,smirnova2019distributionally,derman2020distributional,ho2021partial,badrinath2021robust,curi2021combining,tanabe2022max,goyal2022robust,ding2023seeing,wang2023bring,pmlr-v238-sundhar-ramesh24a}. Robust RL algorithms often employ a minimax formulation, significantly increasing design and implementation complexity. Similar to our approach, \citet{gadot2024bring} propose an algorithm that avoids explicitly solving the minimax problem, instead relying entirely on how trajectories are sampled. Their key result demonstrates that by sampling next states pessimistically with respect to a reward value function, one can derive a robust policy under a rectangular KL-divergence uncertainty set. In addition, \citet{thomas2021safe,as2022constrained,zanger2021safe} use ensembles and pessimism to enforce safety constraints under model uncertainty as done in \algname{SPiDR}. This work differs in that our models are not learned from data but are derived via domain randomization. In addition, we formally show that by only using domain randomization, one can achieve safe transfer to the real system.
Finally, similar to this work, \citet{yu2020mopo} propose to penalize rewards based on model uncertainty. However, their focus is in unconstrained offline RL problems. Model uncertainty is also addressed in other formulations such as risk-sensitive RL \citep{zhang2023soft,kim2023trust}, domain adaptation in RL \citep{chen2024domain}, and curriculum RL \citep{narvekar2020curriculum}. Although these do not focus directly on safety, our work can potentially be adapted to these settings.

\paragraph{Domain randomization.}\looseness=-1Domain randomization enhances policy robustness by training across a variety of environmental scenarios and optimizing average performance. It is widely used in robotics \citep{sadeghi2016cad2rl,tobin2017domain, peng2018sim2real,andrychowicz2020learning}. This work utilizes domain randomization as a key component for robustness, integrating it into the constrained RL framework by adding a robust penalty term to the cost function. Given the limited theoretical analysis on domain randomization \citep{chen2022understandingdomainrandomizationsimtoreal}, we theoretically study it when facing with additional safety requirements, highlighting its limitations and addressing them. Further, similar to our work, \citet{lee2023evaluation} and \citet{kim2024not} combine CMDPs with domain randomization. However, these works apply CMDPs primarily to legged locomotion, using constraints to shape stylistic gait qualities, rather than ensuring safety. Additionally, their evaluations are largely qualitative and assess constraint satisfaction in simulation. Lastly, a comprehensive survey on domain randomization, and other methods for sim-to-real transfer is discussed in more details by \citet{zhao2020sim}.

\end{document}